\documentclass[10pt,twocolumn,letterpaper]{article}

\usepackage[pagenumbers]{cvpr} 

\definecolor{cvprblue}{rgb}{0.21,0.49,0.74}
\usepackage[pagebackref,breaklinks,colorlinks,allcolors=cvprblue]{hyperref}

\def\paperID{} 
\def\confName{CVPR}
\def\confYear{2026}

\usepackage[table]{xcolor} 
\usepackage{tabularx}
\usepackage{array}
\usepackage{booktabs}
\usepackage{multirow}
\usepackage{etoolbox}

\usepackage{placeins}
\usepackage{etoc}
\etocdepthtag.toc{mtchapter}
\etocsettagdepth{mtchapter}{subsection}
\etocsettagdepth{mtappendix}{none}

\definecolor{headergray}{gray}{0.92}
\definecolor{oursblue}{HTML}{E6F0FF}
\newcolumntype{C}{>{\centering\arraybackslash}X}
\newcolumntype{L}{>{\raggedright\arraybackslash}X}

\title{LACON: Training Text-to-Image Model from Uncurated Data}

\author{
Zhiyang Liang\textsuperscript{\rm 1 \rm 2 $\dagger$}
\quad
Ziyu Wan\textsuperscript{\rm 2 $\ast$}
\quad
Hongyu Liu\textsuperscript{\rm 3}
\quad
Dong Chen\textsuperscript{\rm 2}
\\
Qiu Shen\textsuperscript{\rm 1}
\quad
Hao Zhu\textsuperscript{\rm 1}
\quad
Dongdong Chen\textsuperscript{\rm 2}
\vspace{5pt}\\
\textsuperscript{\rm 1}Nanjing University
\qquad
\textsuperscript{\rm 2}Microsoft Research
\qquad
\textsuperscript{\rm 3}HKUST
\vspace{2mm}
\\
Project Page: \href{https://zhiyangliang.github.io/LACON}{LACON}
}

\begin{document}

\twocolumn[{
\renewcommand\twocolumn[1][]{#1}
\maketitle
\begin{center}
    \vspace{-20pt}
    \includegraphics[width=1.0\linewidth]{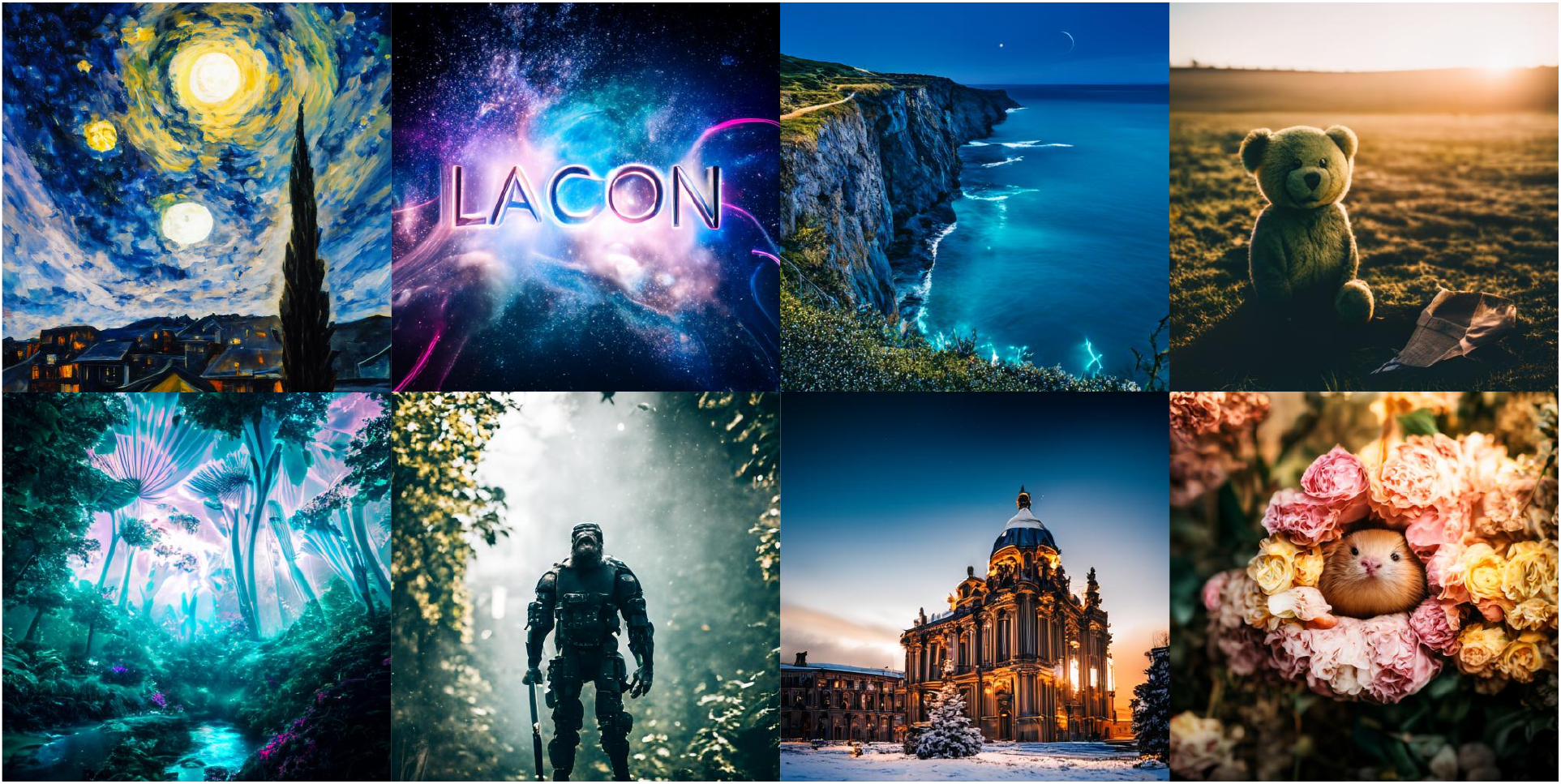}
    \vspace{-20pt}
    \captionsetup{type=figure}
    \caption{Generated images from LACON at a resolution of 512 × 512. 
    }
    \label{fig:teaser}
\end{center}
}]

\maketitle

\renewcommand{\thefootnote}{$\dagger$}
\footnotetext[1]{Work done during the internship at Microsoft Research.}

\renewcommand{\thefootnote}{$\ast$}
\footnotetext[2]{Corresponding author.}

\renewcommand{\thefootnote}{\arabic{footnote}}

\begin{abstract}
The success of modern text-to-image generation is largely attributed to massive, high-quality datasets. Currently, these datasets are curated through a \textit{filter-first} paradigm that aggressively discards low-quality raw data based on the assumption that it is detrimental to model performance. Is the discarded bad data truly useless, or does it hold untapped potential? In this work, we critically re-examine this question. We propose \textbf{LACON} (Labeling-and-Conditioning), a novel training framework that exploits the underlying uncurated data distribution. Instead of filtering, LACON re-purposes quality signals, such as aesthetic scores and watermark probabilities as explicit, quantitative condition labels. The generative model is then trained to learn the full spectrum of data quality, from bad to good. By learning the explicit boundary between high- and low-quality content, LACON achieves superior generation quality compared to baselines trained only on filtered data using the same compute budget, proving the significant value of uncurated data.
\end{abstract}
    
\section{Introduction}
\label{sec:intro}

Large-scale generative foundation models have recently triggered a revolutionary wave of progress across diverse modalities, including image~\cite{baldridge2024imagen, kolors, wan2021high, liu2024playground, batifol2025flux, cai2025hidream}, video~\cite{wan2025wan, zhang2023i2vgen, blattmann2023stable, yang2024cogvideox, kong2024hunyuanvideo, peng2025open}, and 3D asset~\cite{xiang2025structured, li2025triposg, ye2025hi3dgen, wan2024cad, kaplan2020scaling, yan2025object}. Among these, text-to-image (T2I) synthesis has witnessed particularly dramatic advancements. State-of-the-art models, such as Qwen-Image~\cite{wu2025qwen}, GPT-Image-1~\cite{gpt-image-1}, Nano Banana~\cite{nanobanana}, Hunyuan-Image~\cite{cao2025hunyuanimage} and Seedream~\cite{seedream2025seedream}, now generate images with unprecedented visual fidelity, generation diversity, and prompt controllability. These capabilities are fundamentally reshaping entire fields, from content creation and digital art to human-computer interaction.

The success of these foundation models is inseparable from one crucial factor: massive data scale. However, raw scale alone is not the sole determinant of success; the quality and cleanliness of this data are widely considered to be equally, if not more, paramount. It is a widely held assumption that training on bad or uncurated data, such as images with low resolution, poor aesthetics, watermarks, or over-exposure, would tend to degrade model behavior. This very belief has driven the field to adopt a filter-first paradigm, which is not a minor preprocessing step but a core, resource-intensive component in recent SOTA pipelines. For example, Stable Diffusion~\cite{rombach2022high} is not trained on the full LAION-5B dataset \cite{schuhmann2022laion}, but on LAION-Aesthetics, a highly-filtered subset based on aesthetic scores. Qwen-Image \cite{wu2025qwen} proposes a multi-stage filtering pipeline comprising seven sequential stages, while Hunyuan-Image \cite{cao2025hunyuanimage} implemented a comprehensive three-stage process, ultimately discarding over 55\% of its initial raw data.

While this filter-first paradigm has become the de facto standard, we argue that it suffers from two major drawbacks. First, it is conceptually wasteful and statistically inefficient. The vast majority of data (often over 55\%) is simply discarded. Such bad data (e.g., images with low-aesthetic scores, watermarks, or poor text-image alignment) is not mere noise; it represents a rich, unexploited source of information that could potentially improve the underlying knowledge of foundation models. Second, perhaps most critically, filtering creates a fundamental knowledge gap in the model. By being exclusively exposed to a narrow slice of the distribution, the model never learns the full spectrum of data quality. It gains a strong prior for goodness but no explicit understanding of badness, failing to learn the crucial boundary between high-quality and low-quality content. All of these motivate us to ask the following question:  Can we devise a more efficient training method that fully exploits existing data, enabling the model to learn world knowledge, rare concepts, and characteristic low-quality patterns present in so-called bad samples but underrepresented or absent in high-quality data?

In this paper, we propose LACON (Labeling-and-Conditioning), a novel training framework designed to leverage the entire uncurated dataset. Instead of discarding data based on quality signals (e.g., aesthetic scores, text-image alignment, watermark probabilities), LACON re-purposes these signals as explicit, quantitative condition labels $\mathbf{s}$. We then train a conditional text-to-image model $p(\mathbf{x} | \mathbf{y}, \mathbf{s})$ to understand not only the text condition $\mathbf{y}$, but also these fine-grained attribute labels $\mathbf{s}$. This design directly addresses the previously discussed limitations: it is data-rich, utilizing 100\% of the available raw data, and knowledge-complete, as it explicitly teaches the model the full spectrum of data quality, from bad to good. Moreover, LACON provides a principled training-time counterpart to commonly used ad-hoc inference-time heuristics—such as applying classifier-free guidance with negative prompts by explicitly modeling data quality during training. Crucially, it shifts the responsibility of quality control from inference-time guesswork (e.g., tuning negative prompts) to a well-defined training-time specification. As a result, the model no longer needs to be pushed away from failure modes it never encountered; instead, it learns to navigate the quality space directly through the explicit condition $\mathbf{s}$. During inference, we can explicitly control output quality to meet product or application requirements (e.g., aesthetic quality).

\begin{figure*}[!t]
    \centering
    \includegraphics[width=0.9\textwidth]{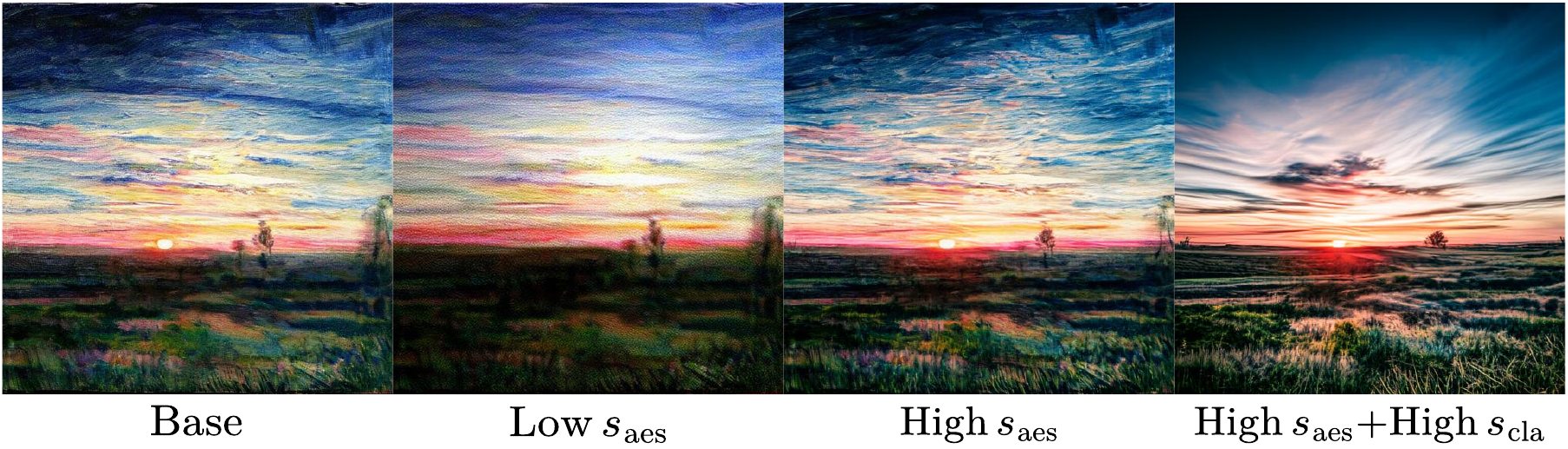}
    \caption{
        \textbf{Comparison of generations produced by the baseline trained on full raw data and by LACON under different conditioning settings}. From left to right: (1) model trained on the full raw dataset without quality conditioning; (2) LACON conditioned on a low aesthetic score $s_{\text{aes}}$; (3) LACON conditioned on a high aesthetic score $s_{\text{aes}}$; (4) LACON jointly conditioned on high aesthetic score $s_{\text{aes}}$ and high clarity score $s_{\text{cla}}$.
    }
    \label{fig: intro_wat}
\end{figure*}

We demonstrate the effectiveness of the LACON framework using a diffusion-based text-to-image generative model through extensive experiments. Our key contributions are threefold:
\begin{itemize}
\item We introduce LACON (Labeling-and-Conditioning), a novel training framework that fundamentally reframes the data curation problem. Instead of the filter-first paradigm, LACON leverages 100\% of the uncurated data by re-purposing quality signals as explicit, learnable conditions.
\item We demonstrate that LACON achieves superior generation quality. By learning the clear boundary between high-quality and low-quality data, our model significantly outperforms baselines trained only on filtered data with the same compute budget, proving the value of bad data.
\item LACON unlocks powerful and quantitative controllability. Our paradigm enables fine-grained control over generation, such as explicitly conditioning on low-quality or high-quality attribute labels as illustrated in Figure~\ref{fig: intro_wat}.
\end{itemize}

\section{Related Work}
\label{sec:related}

\subsection{Text-to-Image Model}
Text-to-image generation has advanced rapidly in recent years, producing visually compelling and semantically aligned outputs. Current approaches largely fall into two dominant paradigms: autoregressive models and diffusion-based models. Autoregressive methods \cite{yu2022scaling,dong2023dreamllm,zhou2024transfusion,wu2025janus} represent images as sequences of discrete tokens and leverage large-scale Transformer architectures to model the joint text–image distribution. This formulation enables strong compositionality and scalability but often struggles with fine-grained spatial coherence due to the sequential nature of generation. In contrast, diffusion-based approaches \cite{nichol2021glide,podell2023sdxl,rombach2022high,gu2022vector,chang2023muse,wei2024enhancing,seedream2025seedream,wu2025qwen} synthesize images by progressively denoising Gaussian noise under text conditioning. Variants such as latent diffusion \cite{rombach2022high} and DiT-style architectures \cite{wei2024enhancing,wu2025qwen} have set new benchmarks for photorealism and high-resolution fidelity. These models excel at capturing rich visual details and global consistency, though they typically require iterative sampling, which can be computationally expensive.
Despite their impressive generative capabilities, both paradigms—especially state-of-the-art systems like \cite{li2024hunyuan,li2024playground,wu2025qwen} depend heavily on massive, high-quality image–text datasets for training. 

\subsection{Data Utilization}
Existing state-of-the-art text-to-image models overwhelmingly adopt a filter-first paradigm, where massive raw datasets are aggressively pruned through multi-stage pipelines before training~\cite{wu2025qwen, gao2025seedream, cao2025hunyuanimage}. For example, Qwen-Image~\cite{wu2025qwen} applies sequential thresholds to eliminate images with low brightness, poor clarity, or artifacts such as watermarks and mosaics. Seedream~\cite{gao2025seedream} goes further by training defect detectors and masking defective regions during gradient computation, while Hunyuan-Image~\cite{cao2025hunyuanimage} employs aesthetic scoring models to enforce unified quality thresholds across all content types. These strategies share a common goal: distill a small, high-quality subset from web-scale corpora to maximize visual fidelity.
However, this paradigm is inherently wasteful and statistically biased. By discarding over half of the raw data, models lose exposure to rare concepts and characteristic low-quality patterns, creating a blind spot in their understanding of the full quality spectrum. In contrast, LACON explores a new strategy: rather than filtering out bad data, we repurpose quality signals as explicit conditioning labels, enabling the model to learn both high-quality and low-quality distributions. This approach transforms data curation from a destructive process into a constructive one, leveraging almost 100\% of available data while preserving controllability.

\subsection{Controllable Generation}
Controllability \cite{zhao2023uni} has long been a central theme in text-to-image synthesis, with prior work introducing adapter modules or conditional branches to steer generation beyond raw text prompts~\cite{rombach2022high}. For instance, IP-Adapter~\cite{ye2023ip} injects image prompts via decoupled cross-attention, Elite~\cite{wei2023elite} encodes user-specific visual concepts into textual embeddings, and VMix~\cite{wu2024vmix} introduces aesthetics adapters for disentangled style control. Different from existing method, LACON enables quality controllability into the training process itself by conditioning on quantitative quality labels from the underlying training data. This design eliminates the need for ad-hoc negative prompts or post-hoc adapters, offering fine-grained, interpretable control over attributes such as watermark level or aesthetic score at inference time.

\section{Method}
\label{sec:method}
This section details our proposed method, LACON (LAbeling-and-CONditioning), for training text-to-image models on uncurated data. We begin in Section~\ref{ssec: method_pre} by providing a brief overview of the prerequisite generative model framework used in our experiments. In Section~\ref{ssec: method_lacon}, we introduce the core components of LACON, detailing our data attribute labeling process, the mechanism for label-conditioning, and the associated training strategy. Finally, we describe the flexible test-time inference strategies enabled by our approach in Section~\ref{ssec: method_infer}.
\begin{figure}[t]
    \centering
    \includegraphics[width=0.5\textwidth]{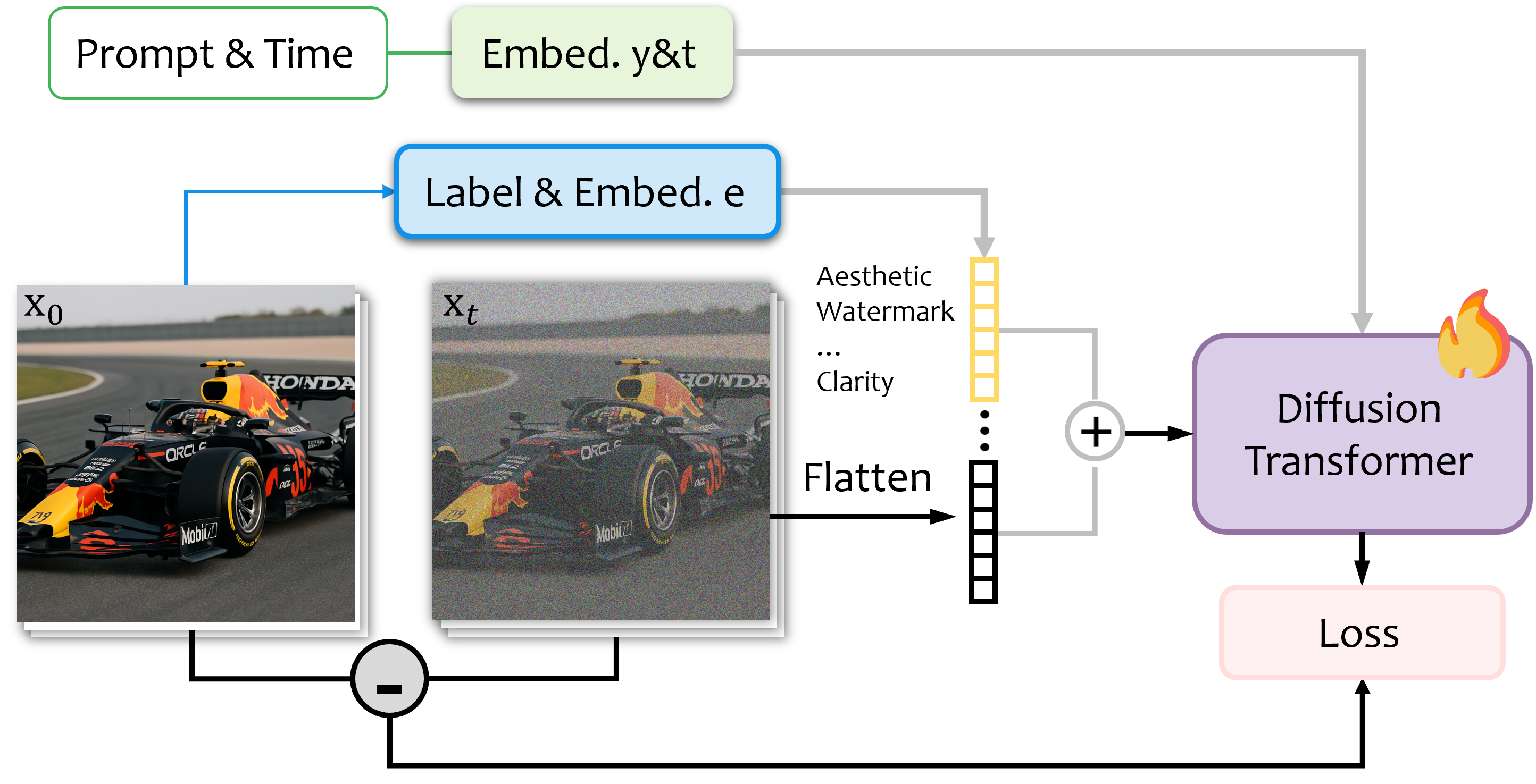}
    \caption{
        $\textbf{Overview of LACON}$. High-level training pipeline that repurposes quality signals as explicit attribute conditioning during training.
    }
    \label{fig: framework}
\end{figure}

\subsection{Preliminaries}
\label{ssec: method_pre}
Text-to-image (T2I) generative models aim to learn the complex conditional probability distribution $p(\mathbf{x}| \mathbf{y})$, where $\mathbf{x}$ is the image and $\mathbf{y}$ is the corresponding text condition. After training, the model enables the sampling of high-quality and diversified visual contents from this distribution given prompts. In the past, most successful T2I models \cite{wu2025qwen,cao2025hunyuanimage} have been trained using filtered high-quality data. In this paper, we propose LACON, a model-agnostic framework that can be generalized to various generative methods~\cite{ramesh2021zero, ho2020denoising, song2020denoising,xie2024sana}, to fully leverage the potential of the entire uncurated dataset.

We demonstrate its efficacy by integrating LACON into a diffusion model. Following standard practice, we adopt the flow matching objective~\cite{lipman2022flow, liu2022flow} with linear interpolation path $\mathbf{x}_t = (1 - t)\mathbf{x} + t\boldsymbol{\epsilon}$, where $\mathbf{x} \sim p(\mathbf{x})$ and $\boldsymbol{\epsilon} \sim \mathcal{N}(\mathbf{0}, \mathbf{I})$, and train the target generative model to predict the velocity $v_{\theta}(\mathbf{x}_t, t, \mathbf{y})=\boldsymbol{\epsilon}-\mathbf{x}$. To ensure the training efficiency, we use Sana~\cite{xie2024sana}, a highly efficient DiT-based~\cite{peebles2023scalable} T2I framework, as our model backbone.

\begin{table*}[t!]
\centering
\caption{Configuration of cluster-centroid anchors for each attribute embedding. The number and placement of anchors ($p_i$) are selected based on the value range of the corresponding quality signals.}
\label{tab:centroid_config}
\resizebox{1.0\textwidth}{!}{%
\begin{tabular}{@{}l p{3cm} c c c l@{}}
\toprule
\textbf{Metric ($s_{\text{metric}}$)} & \textbf{Meaning} & \textbf{Value Range} & \textbf{Clipping} & \textbf{Tokens} ($N$) & \textbf{Centroid Anchors} ($p_i^{\text{metric}}$) \\
\midrule
Aesthetic ($s_{\text{aes}}$)   & \mbox{Aesthetic quality} & 0--10   & None     & 10  & \{0.5, 1.5, \dots, 9.5\} \\
Watermark ($s_{\text{wat}}$)   & \mbox{Watermark probability} & 0--1    & None     & 10 & \{0.05, 0.15, \dots, 0.95\} \\
Clarity ($s_{\text{cla}}$)     & \mbox{Perceptual sharpness} & 0--1M+  & $>3000$  & 10  & \{150, 450, \dots, 2850\} \\
Entropy ($s_{\text{ent}}$)     & \mbox{Information density} & 0--8    & None     & 8  & \{0.5, 1.5, \dots, 7.5\} \\
Luminance ($s_{\text{luma}}$)  & \mbox{Visual brightness} & 0--1    & None     & 10 & \{0.05, 0.15, \dots, 0.95\} \\
\bottomrule
\end{tabular}%
}
\end{table*}

\subsection{LACON}
\label{ssec: method_lacon}
As established in the previous sections, the filter-first paradigm suffers from data waste and creates a knowledge gap inside the model. Instead of viewing quality signals (e.g., aesthetic scores, watermarks) as a pre-training filter, LACON simply re-purposes them as explicit, learnable conditions during training. Technically, this approach reframes the T2I generation task. We shift from the standard objective $v_\theta(\mathbf{x}_t, t, \mathbf{y})$ trained on a filtered dataset, to a multi-conditional objective $v_\theta(\mathbf{x}_t, t, \mathbf{s}, \mathbf{y})$ trained on the entire uncurated dataset. Here, $\mathbf{s}$ is a vector representing the quantitative attribute labels for each sample.

\vspace{0.5em}
\noindent\textbf{Data Attribute Labeling.} Without loss of generality, we consider five condition signals, i.e.,\texttt{aesthetic}, \texttt{watermark}, \texttt{clarity}, \texttt{entropy}, and \texttt{luminance}, which are commonly used in image-filtering pipelines and capture a diverse range of semantic and low-level image properties. Our training corpus consists of 110M images collected from publicly available sources. To implement LACON, we build an automated labeling pipeline that annotates each image with a five-dimensional conditioning vector $\mathbf{s}=[s_{\text{aes}}, s_{\text{wat}}, s_{\text{cla}}, s_{\text{ent}}, s_{\text{luma}}]$. More details and label configurations could be found in Table~\ref{tab:centroid_config}.

\vspace{0.5em}
\noindent\textbf{Condition Embedding.} To leverage the conditional vector $\mathbf{s}$, we must convert them into learnable embeddings that can be ingested by the generative model. We propose a flexible embedding strategy that converts each continuous scalar score into a soft vector representation. We process each of the $K=5$ attributes independently. For the $k$-th attribute (e.g., $s_{\texttt{aes}}$), we define two components: 1) Fixed anchors: A set of $N$ fixed, non-learnable scalar anchor points, $\mathbf{p}^{(k)} = \{p_1^{(k)}, \dots, p_N^{(k)}\}$, where $N$ is chosen based on the value range and granularity of attribute. These anchors partition the continuous space of the attribute, linearly spaced across the expected score range of the specific attribute. 2) Learnable centroids: For each anchor $p_i^{(k)}$, we attach a learnable embedding vector, or centroid token $\mathbf{c}_i^{(k)} \in \mathbb{R}^d$, where the embedding dimension $d$ matches the channel size of the noisy latent $\mathbf{x}_t$.

Given $k$-th attribute $s_{k}$, we compute its affnity scores $u_i^{(k)}$ over different anchor points $p_i^{(k)}$ using a Gaussian Radial Basis Function (RBF) kernel:
$$u_i^{(k)} = \exp\left(-\frac{(s_k - p_i^{(k)})^2}{2\sigma_k^2}\right)$$
Here, $\sigma_k$ is an attribute-specific parameter determined by half the distance between adjacent centroid anchors of the $k$-th attribute, enabling independent smoothness control for each attribute. We then normalize the affinity scores to produce the final weights $w_i^{(k)}$:
$$w_i^{(k)} = \frac{u_i^{(k)}}{\sum_{j=1}^N u_j^{(k)}} $$
The final embedding $\mathbf{e}_k$ for the $k$-th attribute is the weighted sum of all $N$ learnable centroid tokens:
$$\mathbf{e}_k = \sum_{i=1}^N w_i^{(k)} \mathbf{c}_i^{(k)}$$

\vspace{0.5em}
\noindent\textbf{Training.} 
The LACON framework accepts two distinct conditional inputs: the text prompt $\mathbf{y}$ and our new attribute vector $\mathbf{s}$. For prompt $\mathbf{y}$,  we feed its text embedding to the model using cross attention to guide the semantics of the generation. For our new attribute embedding, we adopt a more straightforward strategy, i.e., context conditioning, which simply concatenates the noisy latents $\mathbf{x}_t$ with the condition embedding $\mathbf{e}$ to form a new token sequence as the DiT input so that the model can perceive the complete conditioning context from the very first layer. A high-level illustration of our training framework is provided in Figure~\ref{fig: framework}. We found that this simple concatenation approach works well and effectively conditions the entire denoising process on the desired quality attributes.

\subsection{Inference Strategy}
\label{ssec: method_infer}
LACON's explicit attribute conditioning $\mathbf{s}$ could unlock quantitative control over the generated output to improve the generated visual quality. In this section, we introduce how to do inference with LACON. We still use standard classifier-free guidance (CFG) for the text condition $\mathbf{y}$, but we will introduce two ways: LACON-S and LACON-A to leverage the learned quality attributes.
\begin{itemize}
    \item 
\textit{LACON-S} (standard guidance): The standard and most efficient inference method. It performs CFG on the text condition $\mathbf{y}$ while holding the vector of quality signals $\mathbf{s}$ constant at a single target value (e.g., high aesthetics, no watermark) throughout sampling. 
\item
\textit{LACON-A} (aggressive multi-condition guidance): a more advanced and computationally intensive scheme that applies CFG to both the text condition $\mathbf{y}$ and each quality attribute $s_k$ independently, which allows for dynamic, per-attribute guidance scaling at inference time. At each denoising step $t$, we compute $K+2$ parallel predictions and combine all these guidance vectors by: $$\begin{aligned}
\mathbf{v}_a^t = \mathbf{v}_{\text{base}} & + \omega_c \cdot (\mathbf{v}_{\text{text}} - \mathbf{v}_{\text{base}}) \\
& + \omega_{\text{aes}} \cdot (\mathbf{v}_{\text{aes}} - \mathbf{v}_{\text{text}}) \\
& + \omega_{\text{wat}} \cdot (\mathbf{v}_{\text{wat}} - \mathbf{v}_{\text{text}}) \\
& + \dots \\
& + \omega_{\text{luma}} \cdot (\mathbf{v}_{\text{luma}} - \mathbf{v}_{\text{text}})
\end{aligned}$$
where $\mathbf{v}_{\text{base}}=v_\theta(\mathbf{x}_t, t, \varnothing, \mathbf{s}_{\text{base}})$, $\mathbf{v}_{\text{text}}=v_\theta(\mathbf{x}_t, t, \mathbf{y}, \mathbf{s}_{\text{base}})$. We compute one prediction $\mathbf{v}_\mathbf{s}$ for each of our $K=5$ attributes, where only that single attribute is set to its high-quality target, e.g., $\mathbf{v}_{\text{aes}} = v_\theta(\mathbf{x}_t, t, \mathbf{y}, \mathbf{s}_{\text{high-aes}})$. $\omega_c$ is the text-guidance scale, while $\omega_{\text{aes}}, \dots, \omega_{\text{luma}}$ are individual guidance scales for each quality attribute. This aggressive method provides maximum controllability, allowing a user to dial up or dial down each specific quality metric at inference time to achieve the preferred results.
\end{itemize}

\section{Experiment}
\label{sec:exp}

\begin{figure*}[t!]
    \centering
    \includegraphics[width=1.0\textwidth]{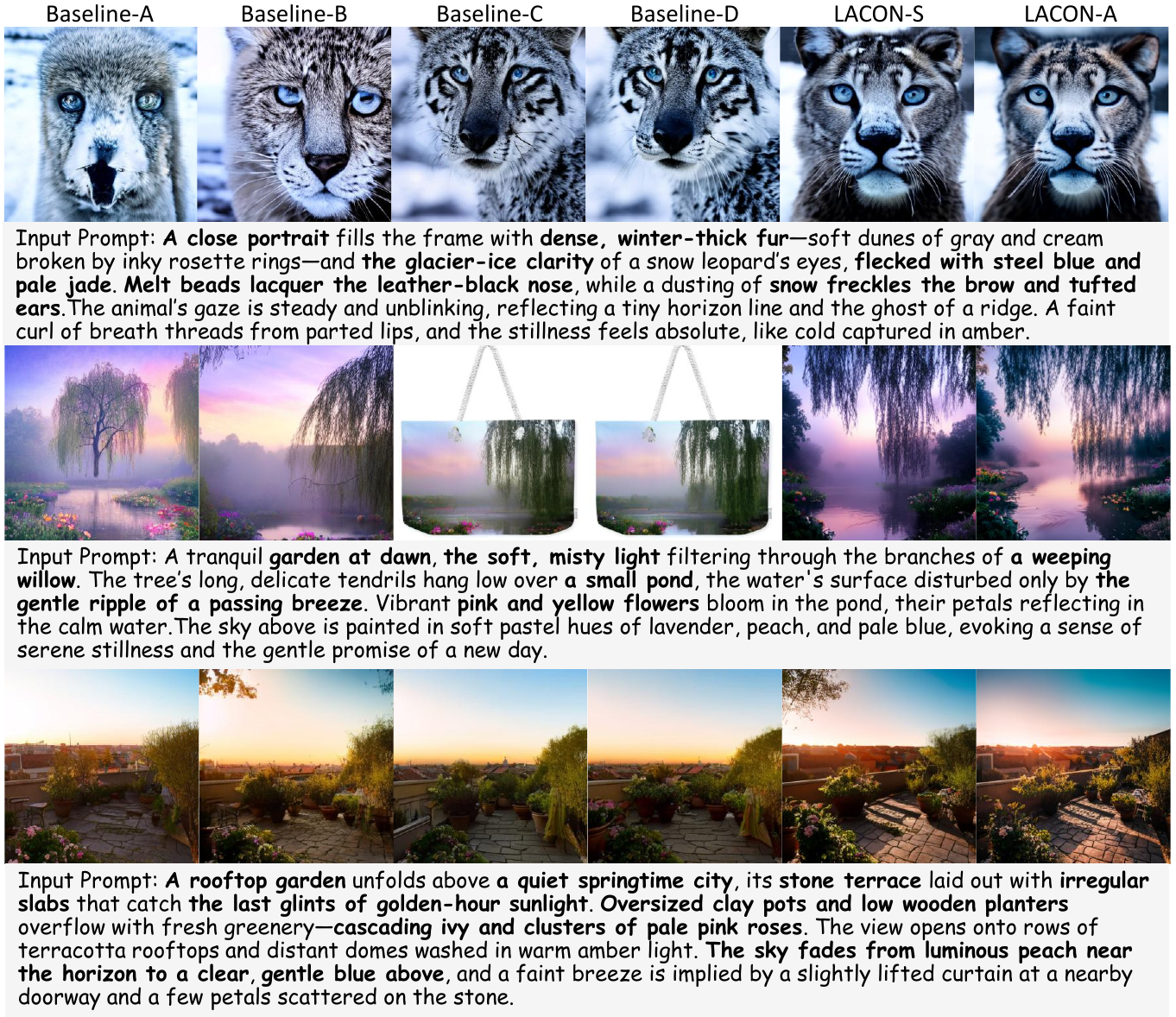}
    \caption{
        Visualization comparison of LACON-S and LACON-A against Baselines, which demonstrate LACON can still achieve superior visual generation quality even when training on the full set images without filtering.
    }
    \label{fig: main_vis}
\end{figure*}

\begin{figure*}[!t]
    \centering
    \includegraphics[width=1.0\textwidth]{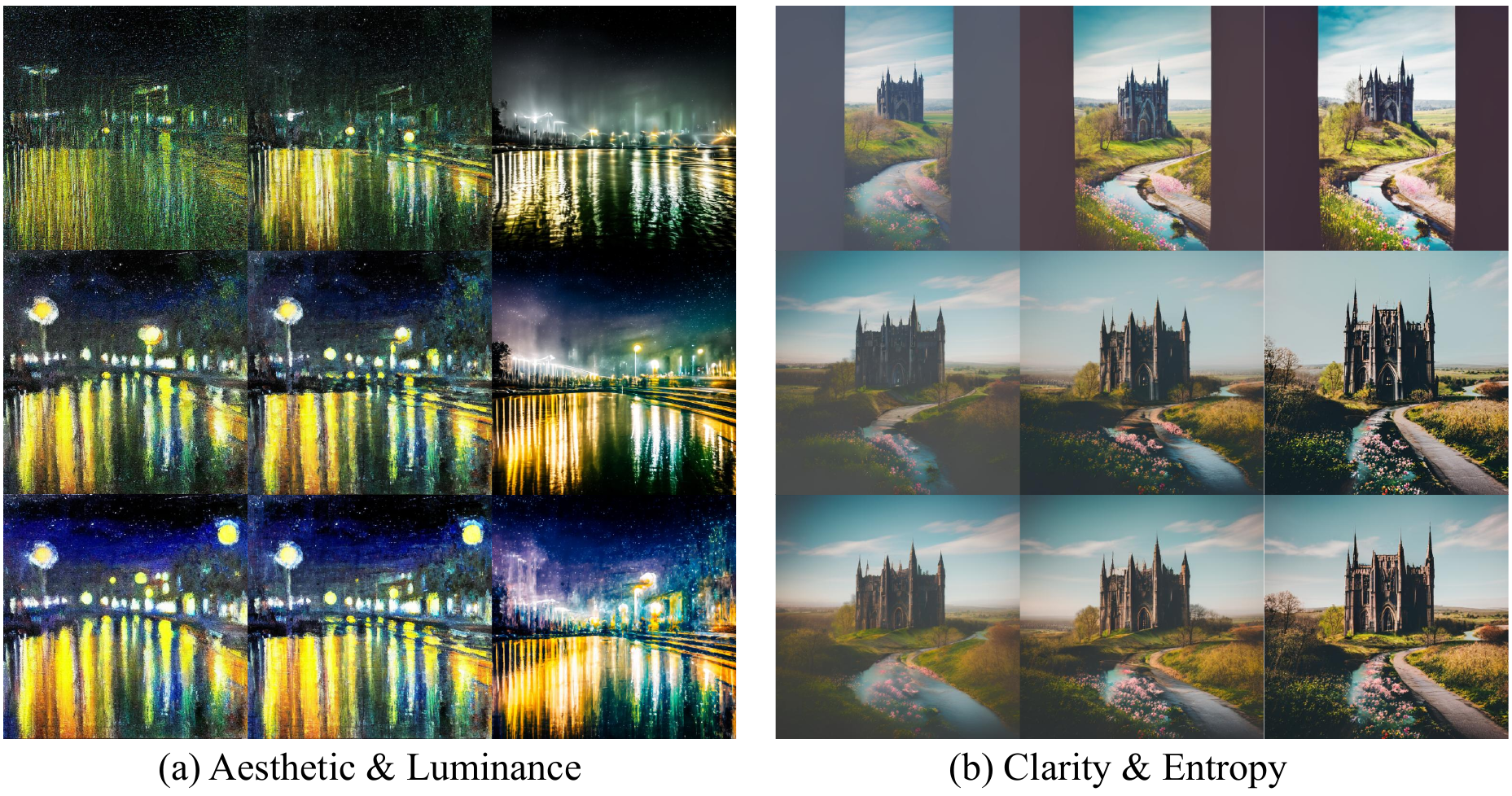}
    \caption{
        (a) \textbf{Comparison of images jointly conditioned on aesthetic score $s_{\mathrm{aes}}$ and HSV-luma score $s_{\mathrm{luma}}$}. From left to right, the columns correspond to $s_{\text{aes}}=3$, $5$, $7$. From top to bottom, the rows correspond to $s_{\text{luma}}=0.3$, $0.4$, $0.5$. (b) \textbf{Comparison of images jointly conditioned on clarity lapvar score $s_{\mathrm{cla}}$ and entropy score $s_{\mathrm{ent}}$}. From left to right, the columns correspond to $s_{\text{cla}}=200$, $1000$, $2500$. From top to bottom, the rows correspond to $s_{\text{ent}}=5$, $6$, $7$.
    }
    \label{fig: joint_control}
\end{figure*}

\subsection{Experiments Setting}
\textbf{Implementation Details.}
Our main experiments are conducted on Sana-0.6B and Sana-1.6B, both pre-trained on 512×512 images. Sana-1.6B is pre-trained for 150K steps with a global batch size of 2048 on 32 NVIDIA A100 GPUs, whereas Sana-0.6B is pre-trained for 100K steps with the same global batch size on 16 NVIDIA A100 GPUs. For high-resolution setting, we further fine-tune the corresponding pre-trained models for an additional 20K steps. All image captions are generated offline using GPT-4o. We report GenEval, DPG, and FID as evaluation metrics. FID is computed on MJHQ-30K, while GenEval and DPG use 533 and 1,065 test prompts, respectively.

\noindent\textbf{Baselines}.
\label{ssec: exp_baselines}
We study how training on subsets constructed using different filtering thresholds affects performance on Sana-1.6B. The thresholds and resulting retained ratios are summarized in Table~\ref{tab:filtering_thresholds}, and the results on GenEval, DPG, and FID are reported in Table~\ref{tab:sana1.6B_ratio_geneval_dpg_fid}. The results reveal a clear trade-off between data quantity and data quality: training on an overly small subset leads to suboptimal performance due to insufficient coverage and diversity. As the retained ratio increases, the training set tends to contain a larger fraction of low-quality samples. Without explicit quality-aware supervision, these samples can degrade model performance.

Based on this sweep, we define four baselines. Baseline-A is trained from scratch on a heavily filtered subset ($\approx 5\%$ of the raw data) obtained with strict quality thresholds; this choice corresponds to the most aggressive filtering setting in our ratio sweep and achieves the worst performance in Table~\ref{tab:sana1.6B_ratio_geneval_dpg_fid}. Baseline-B is trained on a more loosely filtered subset ($\approx 65\%$ of the raw data) under relaxed thresholds; this choice matches the best-performing retention ratio in Table~\ref{tab:sana1.6B_ratio_geneval_dpg_fid} and serves as the strongest filtered-data reference. Baseline-C is trained on the full set of 110M unfiltered images. Baseline-D uses the same training data as Baseline-C, but applies negative-prompt guidance at inference time, which is conceptually related to our LACON strategy of using low-quality signals to explicitly and effectively steer generation toward higher-quality outputs. Specifically, we use the following negative prompt: \textit{low aesthetic score, ugly, bad composition, watermark, logo}.

\begin{table*}[t!]
  \centering
  \small
  \setlength{\tabcolsep}{4pt}
  \renewcommand{\arraystretch}{1.15}
  \caption{The filtering thresholds for different filtering ratios.}
  \label{tab:filtering_thresholds}

  \begin{tabular*}{\textwidth}{@{\extracolsep{\fill}} c|cccccc}
    \specialrule{1.2pt}{0pt}{0pt}
    \textbf{Ratio} & {\normalsize $s_{\texttt{aes-min}}$} & {\normalsize $s_{\texttt{wat-min}}$} & {\normalsize $s_{\texttt{cla-min}}$} & {\normalsize $s_{\texttt{ent-min}}$} & {\normalsize $s_{\texttt{luma-min}}$} & {\normalsize $s_{\texttt{luma-max}}$} \\
    \specialrule{0.6pt}{0pt}{0pt}

    $\approx 5\%$   & 5.0 & 0.3 & 800.0 & 6.0 & 0.1 & 0.9 \\
    $\approx 30\%$  & 4.0 & 0.5 & 600.0 & 4.0 & 0.1 & 0.9 \\
    $\approx 50\%$  & 3.5 & 0.6 & 500.0 & 3.0 & 0.1 & 0.9 \\
    $\approx 65\%$  & 3.0 & 0.7 & 400.0 & 2.0 & 0.1 & 0.9 \\
    $\approx 80\%$  & 3.0 & 0.8 & 200.0 & 2.0 & 0.1 & 0.9 \\
    \specialrule{1.2pt}{0pt}{0pt}
  \end{tabular*}
\end{table*}

\begin{table}[t!]
  \centering
  \small
  \setlength{\tabcolsep}{4pt}
  \renewcommand{\arraystretch}{1.15}
  \caption{Quantitative comparisons on Sana-1.6B are conducted between models trained on subsets obtained under different filtering thresholds, evaluated on GenEval, DPG and FID. The results reveal a clear trade-off between data quantity and data quality in terms of model performance.}
  \label{tab:sana1.6B_ratio_geneval_dpg_fid}

  \begin{tabular*}{\linewidth}{@{\extracolsep{\fill}} c|ccc}
    \specialrule{1.2pt}{0pt}{0pt}
    \multicolumn{1}{c|}{\textbf{Ratio}}
      & \textbf{GenEval}~$\uparrow$ & \textbf{DPG}~$\uparrow$ & \textbf{FID}~$\downarrow$ \\
    \specialrule{0.6pt}{0pt}{0pt}

    $\approx 5\%$   & 58.3 & 73.4 & 14.0 \\
    $\approx 30\%$  & 62.1 & 73.6 & 13.7 \\
    $\approx 50\%$  & 64.5 & 74.5 & 13.4 \\
    $\approx 65\%$  & \textbf{68.0} & \textbf{76.1} & \textbf{13.1} \\
    $\approx 80\%$  & 67.5 & 75.8 & 13.5 \\
    $100\%$         & 67.0 & 75.4 & 13.5 \\
    \specialrule{1.2pt}{0pt}{0pt}
  \end{tabular*}
\end{table}

\begin{table*}[t!]
  \centering
  \small
  \setlength{\tabcolsep}{4pt}
  \renewcommand{\arraystretch}{1.15}
  \caption{Quantitative comparisons on Sana-0.6B and Sana-1.6B, evaluated using GenEval, DPG, and FID, demonstrate that training on the full unfiltered dataset with explicit conditioning enables LACON to consistently outperform the strongest baseline at both 512$\times$512 and 1024$\times$1024 resolutions.}
  \label{tab:sana_geneval_dpg_fid_all}

  \begin{tabular*}{\textwidth}{@{\extracolsep{\fill}} c|ccc|ccc}
    \specialrule{1.2pt}{0pt}{0pt}
    \multicolumn{7}{@{}l@{}}{\textbf{512$\times$512 resolution}} \\
    \specialrule{0.8pt}{0pt}{0pt}

    \multicolumn{1}{c|}{\textbf{Model}}
      & \multicolumn{3}{c|}{\textbf{Sana-0.6B}}
      & \multicolumn{3}{c}{\textbf{Sana-1.6B}} \\
    \specialrule{0.6pt}{0pt}{0pt}

    \multicolumn{1}{c|}{\textbf{Metric}}
      & \textbf{GenEval}~$\uparrow$ & \textbf{DPG}~$\uparrow$ & \textbf{FID}~$\downarrow$
      & \textbf{GenEval}~$\uparrow$ & \textbf{DPG}~$\uparrow$ & \textbf{FID}~$\downarrow$ \\
    \specialrule{0.6pt}{0pt}{0pt}

    Baseline-A              & 55.0 & 68.1 & 14.9 & 58.3 & 73.4 & 14.0 \\
    Baseline-B              & 62.8 & 70.4 & 13.4 & 68.0 & 76.1 & 13.1 \\
    Baseline-C              & 60.8 & 69.7 & 15.4 & 67.0 & 75.4 & 13.5 \\
    Baseline-D              & 61.0 & 69.2 & 15.3 & 67.1 & 75.5 & 13.5 \\
    \specialrule{0.8pt}{0pt}{0pt}
    \textbf{LACON-S}~(Ours)  & 64.4 & \textbf{71.9} & 11.9 & 70.9 & 77.3 & 12.0 \\
    \textbf{LACON-A}~(Ours)  & \textbf{65.6} & 71.8 & \textbf{11.8} & \textbf{71.6} & \textbf{78.1} & \textbf{11.2} \\
    \specialrule{1.2pt}{0pt}{0pt}
  \end{tabular*}

  \begin{tabular*}{\textwidth}{@{\extracolsep{\fill}} c|ccc}
    \multicolumn{4}{@{}l@{}}{\textbf{1024$\times$1024 resolution}} \\
    \specialrule{0.8pt}{0pt}{0pt}

    \multicolumn{1}{c|}{\textbf{Model}}
      & \multicolumn{3}{c}{\textbf{Sana-1.6B}} \\
    \specialrule{0.6pt}{0pt}{0pt}

    \multicolumn{1}{c|}{\textbf{Metric}}
      & \textbf{GenEval}~$\uparrow$ & \textbf{DPG}~$\uparrow$ & \textbf{FID}~$\downarrow$ \\
    \specialrule{0.6pt}{0pt}{0pt}

    Baseline-B              & 68.3 & 76.3 & 12.9 \\
    \specialrule{0.8pt}{0pt}{0pt}
    \textbf{LACON-S}~(Ours)  & 70.9 & 77.6 & 11.4 \\
    \textbf{LACON-A}~(Ours)  & \textbf{71.5} & \textbf{78.8} & \textbf{11.3} \\
    \specialrule{1.2pt}{0pt}{0pt}
  \end{tabular*}
\end{table*}

\subsection{Main Experiments}
\label{ssec: per_comp}
\textbf{Performance Comparison}.
We compare LACON with the baselines described in Section~\ref{ssec: exp_baselines} and report results on GenEval, DPG, and FID for Sana-0.6B and Sana-1.6B in Table~\ref{tab:sana_geneval_dpg_fid_all}. Consistent with the filtering-ratio study, Baseline-B (filtered data) consistently outperforms Baseline-C/D (full unfiltered data), whereas the heavily filtered Baseline-A underperforms, reflecting the expected quality–quantity tension. In contrast, LACON achieves the best performance across GenEval, DPG, and FID for both 512$\times$512 and 1024$\times$1024 resolutions, while being trained on the full raw dataset with explicit quality conditioning. This suggests that samples typically discarded as low quality can still be valuable when their quality is modeled and leveraged rather than removed. LACON-A consistently outperforms LACON-S across most metrics and resolutions, as its stronger guidance yields larger improvements in generation quality and often produces higher-quality visuals.

\noindent
\textbf{Visualization Comparison}.
The visualization comparisons in Figure~\ref{fig: main_vis} further demonstrate that LACON produces higher-quality images while enabling explicit attribute control at inference time. In contrast, training on a heavily filtered subset can create a pronounced knowledge gap, as evidenced by Baseline-A in Figure~\ref{fig: main_vis}. More cases are available in Section~\ref{ssec: knowledge_gap}. Conversely, training directly on the raw dataset may introduce characteristic artifacts at generation time. Beyond inherited watermarks, extremely low-entropy samples can lead to a distinctive failure mode that manifests as blank boundary regions, as observed for Baseline-C/D in Figure~\ref{fig: main_vis}. LACON alleviates both issues while fully leveraging the raw dataset by learning to disentangle high- and low-quality patterns across multiple quality metrics. At inference time, conditioning on the corresponding quality tokens enables the model to suppress artifacts associated with low-entropy samples (e.g., blank boundary regions), resulting in cleaner and more visually coherent outputs.

\subsection{Autoregressive Architecture}
We further evaluate LACON against the strongest baseline (Baseline-B) with Qwen3-0.6B and Qwen3-1.7B using LlamaGen’s VQ-VAE, verifying that our approach generalizes well to autoregressive architectures. Qwen3-1.7B is pre-trained for 120K steps with a global batch size of 2048 on 32 NVIDIA A100 GPUs, while Qwen3-0.6B is pre-trained for 100K steps with a global batch size of 1024 on 16 NVIDIA A100 GPUs. We report results on GenEval, DPG, and FID for both models in Table~\ref{tab:qwen_geneval_dpg_fid}. Across GenEval, DPG, and FID, LACON consistently outperforms Baseline-B on both Qwen3-0.6B and Qwen3-1.7B, with LACON-A further surpassing LACON-S at both scales. These results suggest that LACON’s gains transfer reliably across architectures from diffusion to autoregressive generation.

\begin{table*}[t!]
  \centering
  \small
  \setlength{\tabcolsep}{4pt}
  \renewcommand{\arraystretch}{1.15}
  \caption{Quantitative comparisons on Qwen3-0.6B and Qwen3-1.7B, evaluated using GenEval, DPG, and FID, demonstrate that training on the full unfiltered dataset with explicit conditioning enables LACON to consistently outperform the strongest baseline.}
  \label{tab:qwen_geneval_dpg_fid}

  \begin{tabular*}{\textwidth}{@{\extracolsep{\fill}} c|ccc|ccc}
    \specialrule{1.2pt}{0pt}{0pt}

    \multicolumn{1}{c|}{\textbf{Model}}
      & \multicolumn{3}{c|}{\textbf{Qwen3-0.6B}}
      & \multicolumn{3}{c}{\textbf{Qwen3-1.7B}} \\
    \specialrule{0.6pt}{0pt}{0pt}

    \multicolumn{1}{c|}{\textbf{Metric}}
      & \textbf{GenEval}~$\uparrow$ & \textbf{DPG}~$\uparrow$ & \textbf{FID}~$\downarrow$
      & \textbf{GenEval}~$\uparrow$ & \textbf{DPG}~$\uparrow$ & \textbf{FID}~$\downarrow$ \\
    \specialrule{0.6pt}{0pt}{0pt}

    Baseline-B               & 58.9 & 73.1  & 12.4 & 66.4 & 78.2 & 12.1 \\
    \specialrule{0.8pt}{0pt}{0pt}
    \textbf{LACON-S}~(Ours)   & \textbf{61.3} & 74.9  & 11.4 & 69.7 & 79.4 & 11.0 \\
    \textbf{LACON-A}~(Ours)   & \textbf{61.3} & \textbf{75.4} & \textbf{11.2} & \textbf{70.3} & \textbf{80.1} & \textbf{10.9} \\
    \specialrule{1.2pt}{0pt}{0pt}
  \end{tabular*}
\end{table*}

\begin{figure*}[t!]
    \centering
    \includegraphics[width=0.95\textwidth]{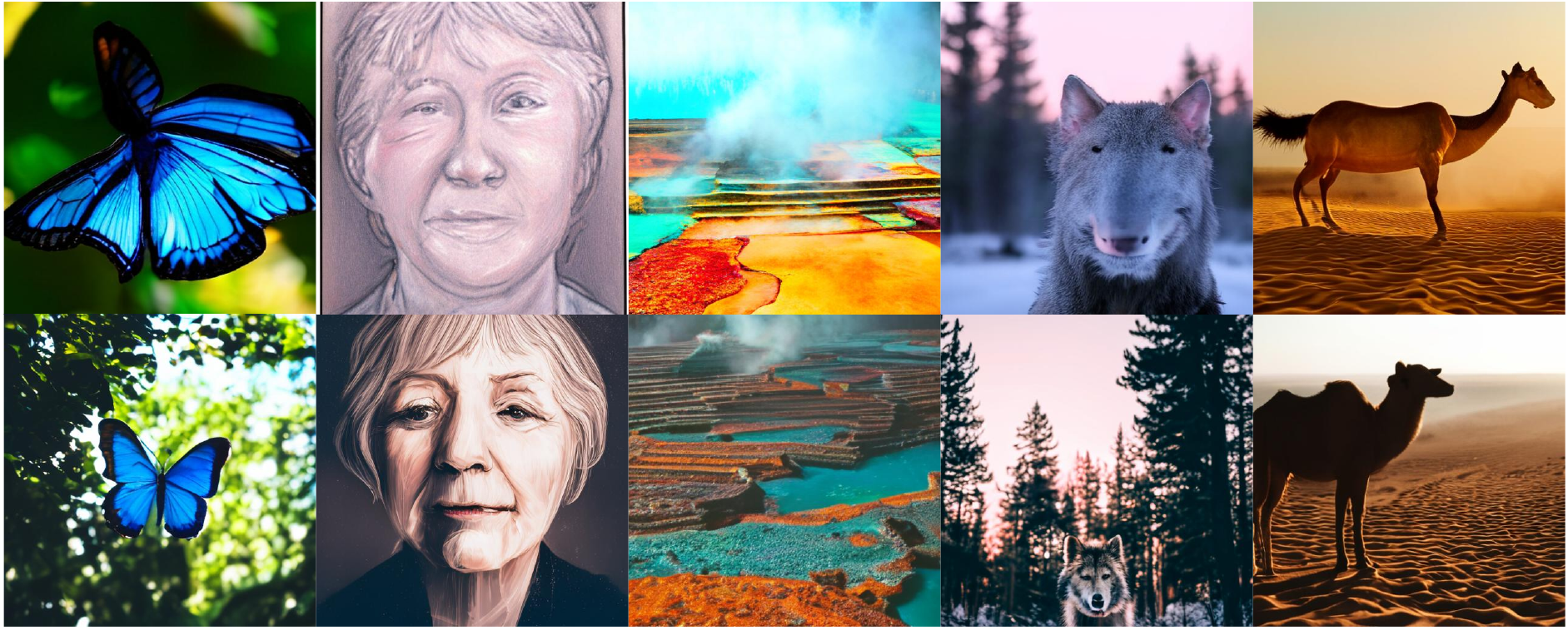}
    \caption{
        The evidence of knowledge gap between Baseline-B (the first row) and LACON (the second row). From left to right, the concepts are about morpho butterfly, pastel portrait of elderly woman, mineral terraces, wolf and camel.
    }
    \label{fig: knowledge_gap}
\end{figure*}

\begin{table}[t!]
  \centering
  \small
  \setlength{\tabcolsep}{4pt}
  \renewcommand{\arraystretch}{1.15}
  \caption{Comparison of different token-injection strategies on Sana-1.6B, including linear interpolation, discrete binning, Fourier-feature embedding, and Gaussian-weighted Cluster Centroids (GCC), evaluated using GenEval, DPG, and FID.}
  \label{tab:token_inject_strategy_results}

  \begin{tabular*}{\linewidth}{@{\extracolsep{\fill}} c|ccc}
    \specialrule{1.2pt}{0pt}{0pt}
    \multicolumn{1}{c|}{\textbf{Method}}
      & \textbf{GenEval}~$\uparrow$ & \textbf{DPG}~$\uparrow$ & \textbf{FID}~$\downarrow$ \\
    \specialrule{0.6pt}{0pt}{0pt}

    Linear Interpolation  & 69.5 & 76.1 & 12.7 \\
    Discrete Binning      & 69.8 & 77.2 & 12.4 \\
    Fourier Feature       & 70.4 & 77.0 & 12.5 \\
    \specialrule{0.8pt}{0pt}{0pt}

    \textbf{GCC}~(Ours)   & \textbf{71.6} & \textbf{78.1} & \textbf{11.2} \\
    \specialrule{1.2pt}{0pt}{0pt}
  \end{tabular*}
\end{table}

\subsection{Knowledeg Gap}
\label{ssec: knowledge_gap}

We evaluate the world-knowledge capabilities of the strongest baseline (Baseline-B) and LACON. Figure~\ref{fig: knowledge_gap} compares images generated for long-tail concepts by Baseline-B and LACON. Although Baseline-B retains approximately 65\% of the full dataset after filtering, it exhibits a noticeable knowledge gap relative to LACON. We attribute this gap to the fact that filtering can discard concept-relevant training samples, thereby reducing coverage of rare or long-tail concepts. These observations suggest a limitation of the filter-first paradigm: samples deemed low-quality can still provide non-trivial, largely untapped supervision that helps models acquire broader world knowledge. By incorporating explicit quality signals, LACON can leverage such long-tail supervision while better separating high- and low-quality content, ultimately improving generation performance.

\subsection{Ablation Studies}
\textbf{Token-injection Strategies}.
We compare four token-injection strategies: linear interpolation, discrete binning, Fourier-feature embedding, and Gaussian-weighted Cluster Centroids (GCC). For linear interpolation, we initialize two learnable tokens per metric at the minimum and maximum of its score range, and obtain each sample’s token by interpolating between the two endpoints according to its score. For discrete binning, we use the same number of tokens per metric as GCC, but quantize scores into bins and assign each sample the token corresponding to its bin. For Fourier-feature embedding, we encode the quality score with Fourier features and map the resulting representation to the token-embedding space via an MLP. GCC is described in detail in Section~\ref{ssec: method_lacon}.

As illustrated in Table~\ref{tab:token_inject_strategy_results}, GCC consistently outperforms the other strategies on GenEval, DPG, and FID. Notably, discrete binning surpasses linear interpolation, indicating that representing each metric with only two learnable tokens and a purely linear parameterization is overly restrictive, limiting the model’s ability to exploit its capacity. In contrast, GCC retains the same token budget as discrete binning but models the score distribution more expressively by softly assigning each sample to multiple anchors. This smooth, overlap-aware encoding captures local structure in the score space and yields more stable, sample-efficient learning from quality-diverse data. Compared with Fourier-feature embedding, GCC is also more robust: Fourier features impose a fixed oscillatory encoding that can be sensitive to small score variations and frequency choices, whereas GCC learns data-adaptive anchors with smooth Gaussian weighting, leading to more stable gradients and better-calibrated conditioning.

\noindent\textbf{Joint Controllability.}
Our LACON framework supports joint conditioning on multiple metric scores, allowing each attribute to vary smoothly from low to high and enabling fine-grained control along different quality dimensions. Figure~\ref{fig: joint_control} (a) illustrates joint control over aesthetic and luminance, while Figure~\ref{fig: joint_control} (b) illustrates joint control over clarity and entropy. For each attribute pair, we vary both attributes across three discrete levels (low/medium/high), resulting in a 3×3 grid with nine combinations. The generated images change smoothly and coherently along both axes, indicating that LACON provides compositional and fine-grained control beyond standard text-only conditioning.

The entropy score reflects image information density. Setting it to a low value at inference time encourages the model to produce low-information outputs, often with blank boundary regions, as illustrated in the first row of Figure~\ref{fig: joint_control} (b). This behavior is also consistent with the blank-boundary patterns observed in the second row of Figure~\ref{fig: main_vis}. Our LACON can mitigate these artifacts by conditioning on quality tokens corresponding to high entropy scores during generation. Similarly, conditioning on a high watermark score effectively suppresses watermark-related patterns. Therefore, although LACON is trained on raw data containing low-entropy or watermarked images, it is still able to produce clean, high-quality outputs at inference time through explicit quality control.

\section{Conclusion}
\label{sec: conclusion}
In this paper, we revisit the prevailing filter-first paradigm in text-to-image generation and propose LACON, a novel training framework that leverages the uncurated data distribution by converting quality signals (e.g., aesthetic scores) into learnable conditioning tokens. This design enables the model to learn the complete spectrum of data quality across multiple dimensions, from bad to good, and to internalize an explicit boundary between them. Across GenEval, DPG and FID, LACON consistently outperforms baselines trained either on filtered high-quality subsets or on full raw data without quality conditioning. These results suggest that data typically discarded as low-quality can be highly valuable when modeled and controlled rather than removed, underscoring the broader potential of uncurated data for building stronger and more controllable generative models.

{
    \small
    \bibliographystyle{ieeenat_fullname}
    \bibliography{main}

\begin{thebibliography}{49}
\providecommand{\natexlab}[1]{#1}
\providecommand{\url}[1]{\texttt{#1}}
\expandafter\ifx\csname urlstyle\endcsname\relax
  \providecommand{\doi}[1]{doi: #1}\else
  \providecommand{\doi}{doi: \begingroup \urlstyle{rm}\Url}\fi

\bibitem[gpt()]{gpt-image-1}
Gpt-image-1.
\newblock \url{https://openai.com/index/image-generation-api/}.

\bibitem[nan()]{nanobanana}
Nano banana.
\newblock \url{https://aistudio.google.com/models/gemini-2-5-flash-image}.

\bibitem[Baldridge et~al.(2024)Baldridge, Bauer, Bhutani, Brichtova, Bunner, Castrejon, Chan, Chen, Dieleman, Du, et~al.]{baldridge2024imagen}
Jason Baldridge, Jakob Bauer, Mukul Bhutani, Nicole Brichtova, Andrew Bunner, Lluis Castrejon, Kelvin Chan, Yichang Chen, Sander Dieleman, Yuqing Du, et~al.
\newblock Imagen 3.
\newblock \emph{arXiv preprint arXiv:2408.07009}, 2024.

\bibitem[Batifol et~al.(2025)Batifol, Blattmann, Boesel, Consul, Diagne, Dockhorn, English, English, Esser, Kulal, et~al.]{batifol2025flux}
Stephen Batifol, Andreas Blattmann, Frederic Boesel, Saksham Consul, Cyril Diagne, Tim Dockhorn, Jack English, Zion English, Patrick Esser, Sumith Kulal, et~al.
\newblock Flux. 1 kontext: Flow matching for in-context image generation and editing in latent space.
\newblock \emph{arXiv e-prints}, pages arXiv--2506, 2025.

\bibitem[Blattmann et~al.(2023)Blattmann, Dockhorn, Kulal, Mendelevitch, Kilian, Lorenz, Levi, English, Voleti, Letts, et~al.]{blattmann2023stable}
Andreas Blattmann, Tim Dockhorn, Sumith Kulal, Daniel Mendelevitch, Maciej Kilian, Dominik Lorenz, Yam Levi, Zion English, Vikram Voleti, Adam Letts, et~al.
\newblock Stable video diffusion: Scaling latent video diffusion models to large datasets.
\newblock \emph{arXiv preprint arXiv:2311.15127}, 2023.

\bibitem[Cai et~al.(2025)Cai, Chen, Chen, Li, Long, Pan, Qiu, Zhang, Gao, Xu, et~al.]{cai2025hidream}
Qi Cai, Jingwen Chen, Yang Chen, Yehao Li, Fuchen Long, Yingwei Pan, Zhaofan Qiu, Yiheng Zhang, Fengbin Gao, Peihan Xu, et~al.
\newblock Hidream-i1: A high-efficient image generative foundation model with sparse diffusion transformer.
\newblock \emph{arXiv preprint arXiv:2505.22705}, 2025.

\bibitem[Cao et~al.(2025)Cao, Chen, Chen, Cheng, Cui, Deng, Dong, Gong, Gu, Gu, et~al.]{cao2025hunyuanimage}
Siyu Cao, Hangting Chen, Peng Chen, Yiji Cheng, Yutao Cui, Xinchi Deng, Ying Dong, Kipper Gong, Tianpeng Gu, Xiusen Gu, et~al.
\newblock Hunyuanimage 3.0 technical report.
\newblock \emph{arXiv preprint arXiv:2509.23951}, 2025.

\bibitem[Chang et~al.(2023)Chang, Zhang, Barber, Maschinot, Lezama, Jiang, Yang, Murphy, Freeman, Rubinstein, et~al.]{chang2023muse}
Huiwen Chang, Han Zhang, Jarred Barber, AJ Maschinot, Jose Lezama, Lu Jiang, Ming-Hsuan Yang, Kevin Murphy, William~T Freeman, Michael Rubinstein, et~al.
\newblock Muse: Text-to-image generation via masked generative transformers.
\newblock \emph{arXiv preprint arXiv:2301.00704}, 2023.

\bibitem[Dong et~al.(2023)Dong, Han, Peng, Qi, Ge, Yang, Zhao, Sun, Zhou, Wei, et~al.]{dong2023dreamllm}
Runpei Dong, Chunrui Han, Yuang Peng, Zekun Qi, Zheng Ge, Jinrong Yang, Liang Zhao, Jianjian Sun, Hongyu Zhou, Haoran Wei, et~al.
\newblock Dreamllm: Synergistic multimodal comprehension and creation.
\newblock \emph{arXiv preprint arXiv:2309.11499}, 2023.

\bibitem[Gao et~al.(2025)Gao, Gong, Guo, Hou, Lai, Li, Li, Lian, Liao, Liu, et~al.]{gao2025seedream}
Yu Gao, Lixue Gong, Qiushan Guo, Xiaoxia Hou, Zhichao Lai, Fanshi Li, Liang Li, Xiaochen Lian, Chao Liao, Liyang Liu, et~al.
\newblock Seedream 3.0 technical report.
\newblock \emph{arXiv preprint arXiv:2504.11346}, 2025.

\bibitem[Gu et~al.(2022)Gu, Chen, Bao, Wen, Zhang, Chen, Yuan, and Guo]{gu2022vector}
Shuyang Gu, Dong Chen, Jianmin Bao, Fang Wen, Bo Zhang, Dongdong Chen, Lu Yuan, and Baining Guo.
\newblock Vector quantized diffusion model for text-to-image synthesis.
\newblock In \emph{Proceedings of the IEEE/CVF conference on computer vision and pattern recognition}, pages 10696--10706, 2022.

\bibitem[Ho et~al.(2020)Ho, Jain, and Abbeel]{ho2020denoising}
Jonathan Ho, Ajay Jain, and Pieter Abbeel.
\newblock Denoising diffusion probabilistic models.
\newblock \emph{Advances in neural information processing systems}, 33:\penalty0 6840--6851, 2020.

\bibitem[Kaplan et~al.(2020)Kaplan, McCandlish, Henighan, Brown, Chess, Child, Gray, Radford, Wu, and Amodei]{kaplan2020scaling}
Jared Kaplan, Sam McCandlish, Tom Henighan, Tom~B Brown, Benjamin Chess, Rewon Child, Scott Gray, Alec Radford, Jeffrey Wu, and Dario Amodei.
\newblock Scaling laws for neural language models.
\newblock \emph{arXiv preprint arXiv:2001.08361}, 2020.

\bibitem[Kong et~al.(2024)Kong, Tian, Zhang, Min, Dai, Zhou, Xiong, Li, Wu, Zhang, et~al.]{kong2024hunyuanvideo}
Weijie Kong, Qi Tian, Zijian Zhang, Rox Min, Zuozhuo Dai, Jin Zhou, Jiangfeng Xiong, Xin Li, Bo Wu, Jianwei Zhang, et~al.
\newblock Hunyuanvideo: A systematic framework for large video generative models.
\newblock \emph{arXiv preprint arXiv:2412.03603}, 2024.

\bibitem[Li et~al.(2024{\natexlab{a}})Li, Kamko, Akhgari, Sabet, Xu, and Doshi]{li2024playground}
Daiqing Li, Aleks Kamko, Ehsan Akhgari, Ali Sabet, Linmiao Xu, and Suhail Doshi.
\newblock Playground v2. 5: Three insights towards enhancing aesthetic quality in text-to-image generation.
\newblock \emph{arXiv preprint arXiv:2402.17245}, 2024{\natexlab{a}}.

\bibitem[Li et~al.(2025)Li, Zou, Liu, Wang, Liang, Yu, Liu, Guo, Liang, Ouyang, et~al.]{li2025triposg}
Yangguang Li, Zi-Xin Zou, Zexiang Liu, Dehu Wang, Yuan Liang, Zhipeng Yu, Xingchao Liu, Yuan-Chen Guo, Ding Liang, Wanli Ouyang, et~al.
\newblock Triposg: High-fidelity 3d shape synthesis using large-scale rectified flow models.
\newblock \emph{arXiv preprint arXiv:2502.06608}, 2025.

\bibitem[Li et~al.(2024{\natexlab{b}})Li, Zhang, Lin, Xiong, Long, Deng, Zhang, Liu, Huang, Xiao, et~al.]{li2024hunyuan}
Zhimin Li, Jianwei Zhang, Qin Lin, Jiangfeng Xiong, Yanxin Long, Xinchi Deng, Yingfang Zhang, Xingchao Liu, Minbin Huang, Zedong Xiao, et~al.
\newblock Hunyuan-dit: A powerful multi-resolution diffusion transformer with fine-grained chinese understanding.
\newblock \emph{arXiv preprint arXiv:2405.08748}, 2024{\natexlab{b}}.

\bibitem[Lipman et~al.(2022)Lipman, Chen, Ben-Hamu, Nickel, and Le]{lipman2022flow}
Yaron Lipman, Ricky~TQ Chen, Heli Ben-Hamu, Maximilian Nickel, and Matt Le.
\newblock Flow matching for generative modeling.
\newblock \emph{arXiv preprint arXiv:2210.02747}, 2022.

\bibitem[Liu et~al.(2024)Liu, Akhgari, Visheratin, Kamko, Xu, Shrirao, Lambert, Souza, Doshi, and Li]{liu2024playground}
Bingchen Liu, Ehsan Akhgari, Alexander Visheratin, Aleks Kamko, Linmiao Xu, Shivam Shrirao, Chase Lambert, Joao Souza, Suhail Doshi, and Daiqing Li.
\newblock Playground v3: Improving text-to-image alignment with deep-fusion large language models.
\newblock \emph{arXiv preprint arXiv:2409.10695}, 2024.

\bibitem[Liu et~al.(2022)Liu, Gong, and Liu]{liu2022flow}
Xingchao Liu, Chengyue Gong, and Qiang Liu.
\newblock Flow straight and fast: Learning to generate and transfer data with rectified flow.
\newblock \emph{arXiv preprint arXiv:2209.03003}, 2022.

\bibitem[Nichol et~al.(2021)Nichol, Dhariwal, Ramesh, Shyam, Mishkin, McGrew, Sutskever, and Chen]{nichol2021glide}
Alex Nichol, Prafulla Dhariwal, Aditya Ramesh, Pranav Shyam, Pamela Mishkin, Bob McGrew, Ilya Sutskever, and Mark Chen.
\newblock Glide: Towards photorealistic image generation and editing with text-guided diffusion models.
\newblock \emph{arXiv preprint arXiv:2112.10741}, 2021.

\bibitem[Peebles and Xie(2023)]{peebles2023scalable}
William Peebles and Saining Xie.
\newblock Scalable diffusion models with transformers.
\newblock In \emph{Proceedings of the IEEE/CVF international conference on computer vision}, pages 4195--4205, 2023.

\bibitem[Peng et~al.(2025)Peng, Zheng, Shen, Young, Guo, Wang, Xu, Liu, Jiang, Li, et~al.]{peng2025open}
Xiangyu Peng, Zangwei Zheng, Chenhui Shen, Tom Young, Xinying Guo, Binluo Wang, Hang Xu, Hongxin Liu, Mingyan Jiang, Wenjun Li, et~al.
\newblock Open-sora 2.0: Training a commercial-level video generation model in $\$$200 k.
\newblock \emph{arXiv preprint arXiv:2503.09642}, 2025.

\bibitem[Podell et~al.(2023)Podell, English, Lacey, Blattmann, Dockhorn, M{\"u}ller, Penna, and Rombach]{podell2023sdxl}
Dustin Podell, Zion English, Kyle Lacey, Andreas Blattmann, Tim Dockhorn, Jonas M{\"u}ller, Joe Penna, and Robin Rombach.
\newblock Sdxl: Improving latent diffusion models for high-resolution image synthesis.
\newblock \emph{arXiv preprint arXiv:2307.01952}, 2023.

\bibitem[Ramesh et~al.(2021)Ramesh, Pavlov, Goh, Gray, Voss, Radford, Chen, and Sutskever]{ramesh2021zero}
Aditya Ramesh, Mikhail Pavlov, Gabriel Goh, Scott Gray, Chelsea Voss, Alec Radford, Mark Chen, and Ilya Sutskever.
\newblock Zero-shot text-to-image generation.
\newblock In \emph{International conference on machine learning}, pages 8821--8831. Pmlr, 2021.

\bibitem[Rombach et~al.(2022)Rombach, Blattmann, Lorenz, Esser, and Ommer]{rombach2022high}
Robin Rombach, Andreas Blattmann, Dominik Lorenz, Patrick Esser, and Bj{\"o}rn Ommer.
\newblock High-resolution image synthesis with latent diffusion models.
\newblock In \emph{Proceedings of the IEEE/CVF conference on computer vision and pattern recognition}, pages 10684--10695, 2022.

\bibitem[Schuhmann et~al.(2022)Schuhmann, Beaumont, Vencu, Gordon, Wightman, Cherti, Coombes, Katta, Mullis, Wortsman, et~al.]{schuhmann2022laion}
Christoph Schuhmann, Romain Beaumont, Richard Vencu, Cade Gordon, Ross Wightman, Mehdi Cherti, Theo Coombes, Aarush Katta, Clayton Mullis, Mitchell Wortsman, et~al.
\newblock Laion-5b: An open large-scale dataset for training next generation image-text models.
\newblock \emph{Advances in neural information processing systems}, 35:\penalty0 25278--25294, 2022.

\bibitem[Seedream et~al.(2025)Seedream, Chen, Gao, Gong, Guo, Guo, Guo, Hou, Huang, Huang, et~al.]{seedream2025seedream}
Team Seedream, Yunpeng Chen, Yu Gao, Lixue Gong, Meng Guo, Qiushan Guo, Zhiyao Guo, Xiaoxia Hou, Weilin Huang, Yixuan Huang, et~al.
\newblock Seedream 4.0: Toward next-generation multimodal image generation.
\newblock \emph{arXiv preprint arXiv:2509.20427}, 2025.

\bibitem[Song et~al.(2020)Song, Meng, and Ermon]{song2020denoising}
Jiaming Song, Chenlin Meng, and Stefano Ermon.
\newblock Denoising diffusion implicit models.
\newblock \emph{arXiv preprint arXiv:2010.02502}, 2020.

\bibitem[Team(2024)]{kolors}
Kolors Team.
\newblock Kolors: Effective training of diffusion model for photorealistic text-to-image synthesis.
\newblock \emph{arXiv preprint}, 2024.

\bibitem[Tschannen et~al.(2025)Tschannen, Gritsenko, Wang, Naeem, Alabdulmohsin, Parthasarathy, Evans, Beyer, Xia, Mustafa, et~al.]{tschannen2025siglip}
Michael Tschannen, Alexey Gritsenko, Xiao Wang, Muhammad~Ferjad Naeem, Ibrahim Alabdulmohsin, Nikhil Parthasarathy, Talfan Evans, Lucas Beyer, Ye Xia, Basil Mustafa, et~al.
\newblock Siglip 2: Multilingual vision-language encoders with improved semantic understanding, localization, and dense features.
\newblock \emph{arXiv preprint arXiv:2502.14786}, 2025.

\bibitem[Wan et~al.(2025)Wan, Wang, Ai, Wen, Mao, Xie, Chen, Yu, Zhao, Yang, et~al.]{wan2025wan}
Team Wan, Ang Wang, Baole Ai, Bin Wen, Chaojie Mao, Chen-Wei Xie, Di Chen, Feiwu Yu, Haiming Zhao, Jianxiao Yang, et~al.
\newblock Wan: Open and advanced large-scale video generative models.
\newblock \emph{arXiv preprint arXiv:2503.20314}, 2025.

\bibitem[Wan et~al.(2021)Wan, Zhang, Chen, and Liao]{wan2021high}
Ziyu Wan, Jingbo Zhang, Dongdong Chen, and Jing Liao.
\newblock High-fidelity pluralistic image completion with transformers.
\newblock In \emph{Proceedings of the IEEE/CVF international conference on computer vision}, pages 4692--4701, 2021.

\bibitem[Wan et~al.(2024)Wan, Paschalidou, Huang, Liu, Shen, Xiang, Liao, and Guibas]{wan2024cad}
Ziyu Wan, Despoina Paschalidou, Ian Huang, Hongyu Liu, Bokui Shen, Xiaoyu Xiang, Jing Liao, and Leonidas Guibas.
\newblock Cad: Photorealistic 3d generation via adversarial distillation.
\newblock In \emph{Proceedings of the IEEE/CVF Conference on Computer Vision and Pattern Recognition}, pages 10194--10207, 2024.

\bibitem[Wei et~al.(2024)Wei, Chen, Zhou, and Pan]{wei2024enhancing}
Tianyi Wei, Dongdong Chen, Yifan Zhou, and Xingang Pan.
\newblock Enhancing mmdit-based text-to-image models for similar subject generation.
\newblock \emph{arXiv preprint arXiv:2411.18301}, 2024.

\bibitem[Wei et~al.(2023)Wei, Zhang, Ji, Bai, Zhang, and Zuo]{wei2023elite}
Yuxiang Wei, Yabo Zhang, Zhilong Ji, Jinfeng Bai, Lei Zhang, and Wangmeng Zuo.
\newblock Elite: Encoding visual concepts into textual embeddings for customized text-to-image generation.
\newblock In \emph{Proceedings of the IEEE/CVF International Conference on Computer Vision}, pages 15943--15953, 2023.

\bibitem[Wu et~al.(2025{\natexlab{a}})Wu, Chen, Wu, Ma, Liu, Pan, Liu, Xie, Yu, Ruan, et~al.]{wu2025janus}
Chengyue Wu, Xiaokang Chen, Zhiyu Wu, Yiyang Ma, Xingchao Liu, Zizheng Pan, Wen Liu, Zhenda Xie, Xingkai Yu, Chong Ruan, et~al.
\newblock Janus: Decoupling visual encoding for unified multimodal understanding and generation.
\newblock In \emph{Proceedings of the Computer Vision and Pattern Recognition Conference}, pages 12966--12977, 2025{\natexlab{a}}.

\bibitem[Wu et~al.(2025{\natexlab{b}})Wu, Li, Zhou, Lin, Gao, Yan, Yin, Bai, Xu, Chen, et~al.]{wu2025qwen}
Chenfei Wu, Jiahao Li, Jingren Zhou, Junyang Lin, Kaiyuan Gao, Kun Yan, Sheng-ming Yin, Shuai Bai, Xiao Xu, Yilei Chen, et~al.
\newblock Qwen-image technical report.
\newblock \emph{arXiv preprint arXiv:2508.02324}, 2025{\natexlab{b}}.

\bibitem[Wu et~al.(2024)Wu, Ding, Huang, Liu, and He]{wu2024vmix}
Shaojin Wu, Fei Ding, Mengqi Huang, Wei Liu, and Qian He.
\newblock Vmix: Improving text-to-image diffusion model with cross-attention mixing control.
\newblock \emph{arXiv preprint arXiv:2412.20800}, 2024.

\bibitem[Xiang et~al.(2025)Xiang, Lv, Xu, Deng, Wang, Zhang, Chen, Tong, and Yang]{xiang2025structured}
Jianfeng Xiang, Zelong Lv, Sicheng Xu, Yu Deng, Ruicheng Wang, Bowen Zhang, Dong Chen, Xin Tong, and Jiaolong Yang.
\newblock Structured 3d latents for scalable and versatile 3d generation.
\newblock In \emph{Proceedings of the Computer Vision and Pattern Recognition Conference}, pages 21469--21480, 2025.

\bibitem[Xie et~al.(2024)Xie, Chen, Chen, Cai, Tang, Lin, Zhang, Li, Zhu, Lu, et~al.]{xie2024sana}
Enze Xie, Junsong Chen, Junyu Chen, Han Cai, Haotian Tang, Yujun Lin, Zhekai Zhang, Muyang Li, Ligeng Zhu, Yao Lu, et~al.
\newblock Sana: Efficient high-resolution image synthesis with linear diffusion transformers.
\newblock \emph{arXiv preprint arXiv:2410.10629}, 2024.

\bibitem[Yan et~al.(2025)Yan, Lee, Wan, and Chang]{yan2025object}
Xingguang Yan, Han-Hung Lee, Ziyu Wan, and Angel~X Chang.
\newblock An object is worth 64$\times$ 64 pixels: Generating 3d object via image diffusion.
\newblock In \emph{2025 International Conference on 3D Vision (3DV)}, pages 123--133. IEEE, 2025.

\bibitem[Yang et~al.(2024)Yang, Teng, Zheng, Ding, Huang, Xu, Yang, Hong, Zhang, Feng, et~al.]{yang2024cogvideox}
Zhuoyi Yang, Jiayan Teng, Wendi Zheng, Ming Ding, Shiyu Huang, Jiazheng Xu, Yuanming Yang, Wenyi Hong, Xiaohan Zhang, Guanyu Feng, et~al.
\newblock Cogvideox: Text-to-video diffusion models with an expert transformer.
\newblock \emph{arXiv preprint arXiv:2408.06072}, 2024.

\bibitem[Ye et~al.(2025)Ye, Wu, Lu, Chang, Guo, Zhou, Zhao, and Han]{ye2025hi3dgen}
Chongjie Ye, Yushuang Wu, Ziteng Lu, Jiahao Chang, Xiaoyang Guo, Jiaqing Zhou, Hao Zhao, and Xiaoguang Han.
\newblock Hi3dgen: High-fidelity 3d geometry generation from images via normal bridging.
\newblock \emph{arXiv preprint arXiv:2503.22236}, 3:\penalty0 2, 2025.

\bibitem[Ye et~al.(2023)Ye, Zhang, Liu, Han, and Yang]{ye2023ip}
Hu Ye, Jun Zhang, Sibo Liu, Xiao Han, and Wei Yang.
\newblock Ip-adapter: Text compatible image prompt adapter for text-to-image diffusion models.
\newblock \emph{arXiv preprint arXiv:2308.06721}, 2023.

\bibitem[Yu et~al.(2022)Yu, Xu, Koh, Luong, Baid, Wang, Vasudevan, Ku, Yang, Ayan, et~al.]{yu2022scaling}
Jiahui Yu, Yuanzhong Xu, Jing~Yu Koh, Thang Luong, Gunjan Baid, Zirui Wang, Vijay Vasudevan, Alexander Ku, Yinfei Yang, Burcu~Karagol Ayan, et~al.
\newblock Scaling autoregressive models for content-rich text-to-image generation.
\newblock \emph{arXiv preprint arXiv:2206.10789}, 2\penalty0 (3):\penalty0 5, 2022.

\bibitem[Zhang et~al.(2023)Zhang, Wang, Zhang, Zhao, Yuan, Qin, Wang, Zhao, and Zhou]{zhang2023i2vgen}
Shiwei Zhang, Jiayu Wang, Yingya Zhang, Kang Zhao, Hangjie Yuan, Zhiwu Qin, Xiang Wang, Deli Zhao, and Jingren Zhou.
\newblock I2vgen-xl: High-quality image-to-video synthesis via cascaded diffusion models.
\newblock \emph{arXiv preprint arXiv:2311.04145}, 2023.

\bibitem[Zhao et~al.(2023)Zhao, Chen, Chen, Bao, Hao, Yuan, and Wong]{zhao2023uni}
Shihao Zhao, Dongdong Chen, Yen-Chun Chen, Jianmin Bao, Shaozhe Hao, Lu Yuan, and Kwan-Yee~K Wong.
\newblock Uni-controlnet: All-in-one control to text-to-image diffusion models.
\newblock \emph{Advances in Neural Information Processing Systems}, 36:\penalty0 11127--11150, 2023.

\bibitem[Zhou et~al.(2024)Zhou, Yu, Babu, Tirumala, Yasunaga, Shamis, Kahn, Ma, Zettlemoyer, and Levy]{zhou2024transfusion}
Chunting Zhou, Lili Yu, Arun Babu, Kushal Tirumala, Michihiro Yasunaga, Leonid Shamis, Jacob Kahn, Xuezhe Ma, Luke Zettlemoyer, and Omer Levy.
\newblock Transfusion: Predict the next token and diffuse images with one multi-modal model.
\newblock \emph{arXiv preprint arXiv:2408.11039}, 2024.

\end{thebibliography}
}

\appendix 
\clearpage

\etocdepthtag.toc{mtappendix}
\etocsettagdepth{mtchapter}{none}
\etocsettagdepth{mtappendix}{section}
\setcounter{page}{1}

\newcommand{\TOCLeader}{\nobreak\leaders\hbox to 0.45em{\hss.\hss}\hfill\nobreak}
\newcommand{\TOCPage}[1]{\makebox[1.7em][r]{#1}}

\renewcommand{\contentsname}{}
\etocsetstyle{section}
  {}
  {}
  {%
    \par\noindent
    \Large
    \makebox[1.7em][l]{\etocnumber}%
    \etocname
    \TOCLeader
    \TOCPage{\etocpage}\par
    \vspace{2em}
  }
  {}

\maketitlesupplementary
\onecolumn

\FloatBarrier
\section{Visualization of Controllability}
\FloatBarrier

\begin{figure}
    \centering
    \vspace{-3mm}
    \includegraphics[width=0.8\textwidth]{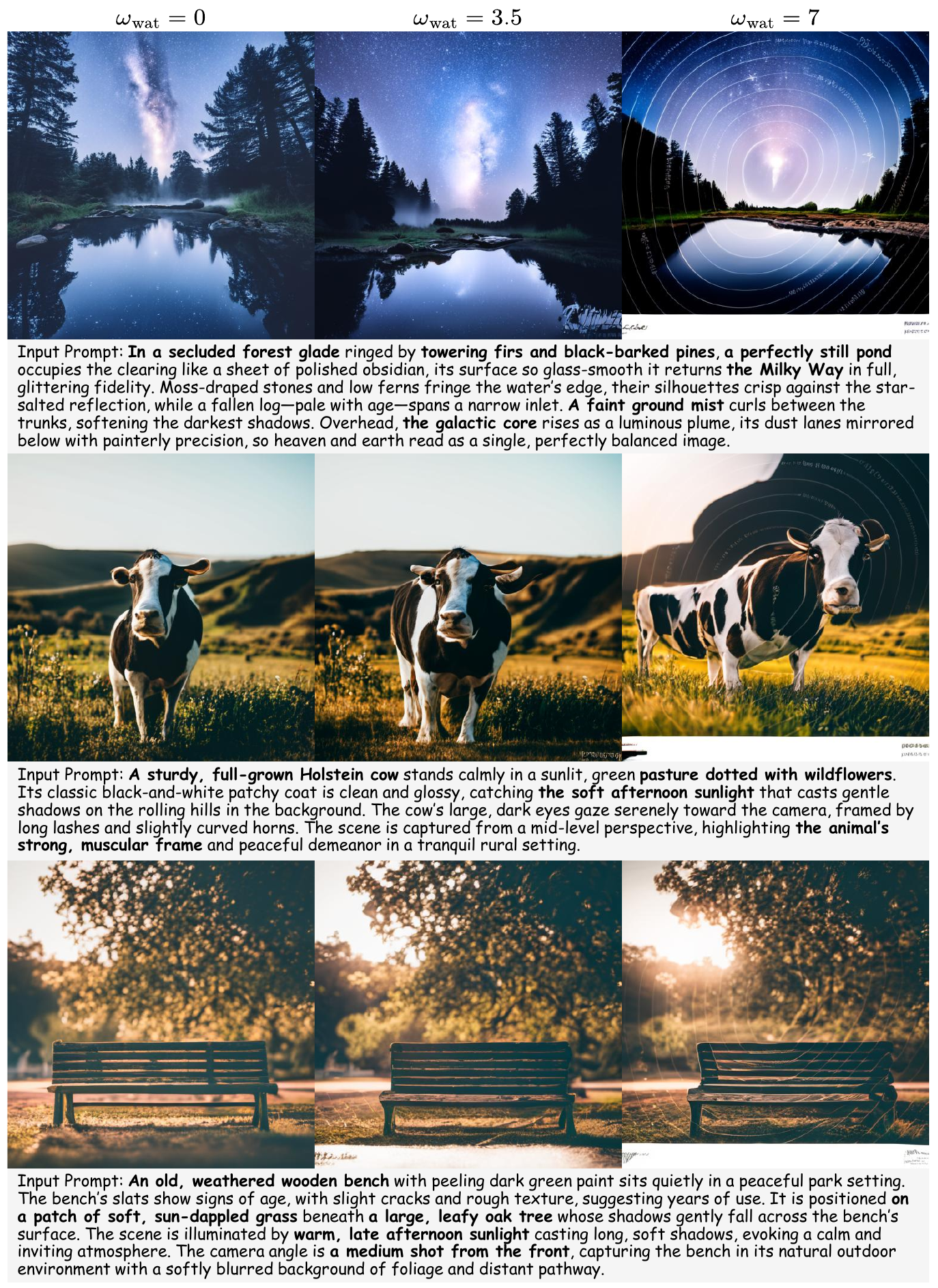}
    \caption{
        \textbf{Visualization comparison of images conditioned on watermark guidance scale $\omega_{\text{wat}}$}. From left to right, the columns correspond to $\omega_{\text{wat}}=0$, $3.5$, $7$ with $s_{\text{wat-low}}=0.05$ and $s_{\text{wat-high}}=0.95$.
    }
    \label{fig: token_wat}
\end{figure}

\begin{figure*}[htbp]
    \centering
    \includegraphics[width=0.9\textwidth]{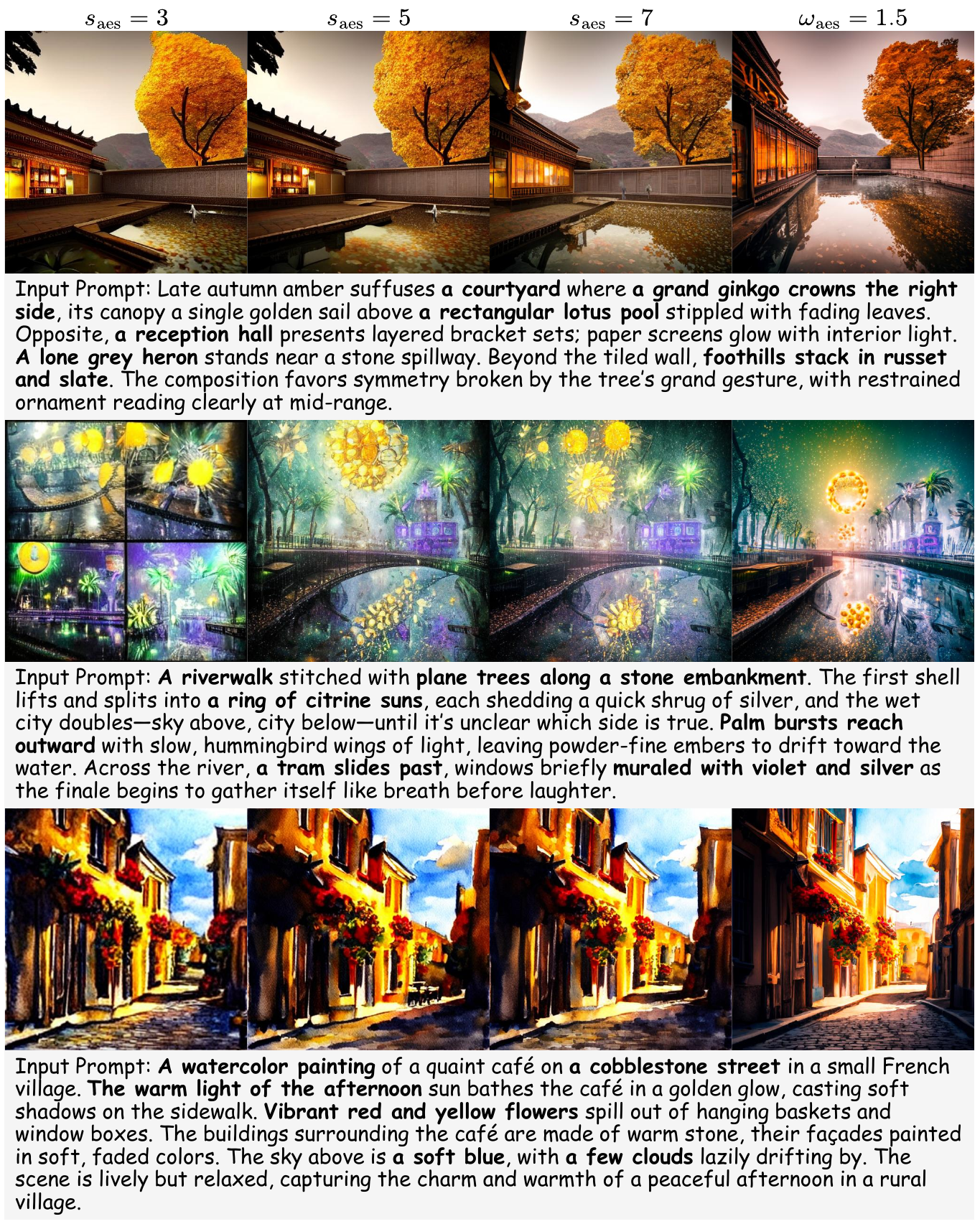}
    \caption{\textbf{Visualization comparison of images conditioned on aesthetic score $s_{\text{aes}}$ and aesthetic guidance scale $\omega_{\text{aes}}$}. From left to right, the columns correspond to $s_{\text{aes}}=3$, $5$, $7$, and $\omega_{\text{aes}}=1.5$ with $s_{\text{aes-low}}=5$ and $s_{\text{aes-high}}=7$.}
    \label{fig: token_aes}
\end{figure*}

\begin{figure*}[htbp]
    \centering
    \includegraphics[width=0.9\textwidth]{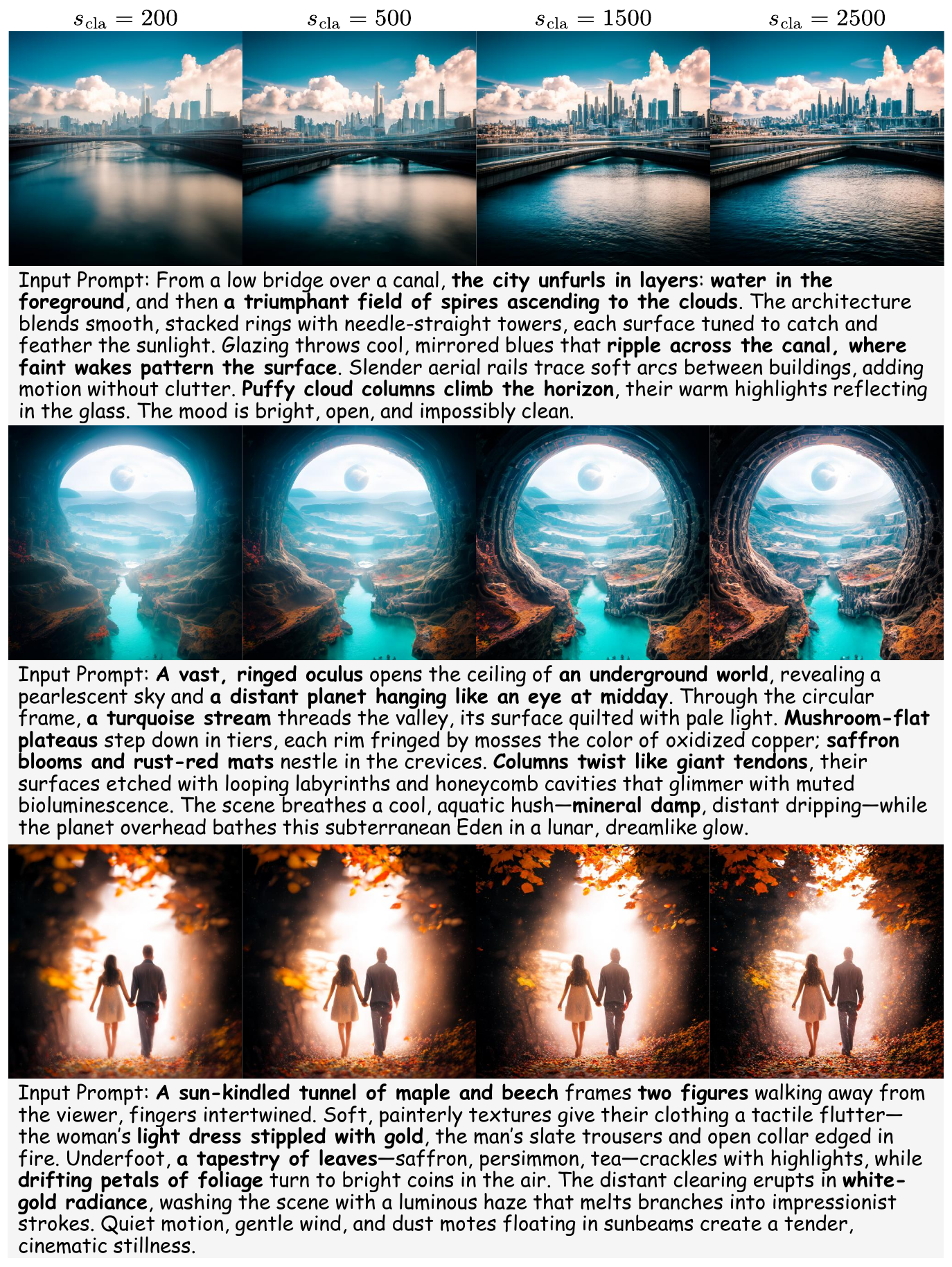}
    \caption{
        \textbf{Visualization comparison of images conditioned on clarity score $s_{\text{cla}}$}. From left to right, the columns correspond to $s_{\text{cla}}=200$, $500$, $1500$, $2500$, respectively.
    }
    \label{fig: token_cla}
\end{figure*}

\begin{figure*}[htbp]
    \centering
    \includegraphics[width=0.9\textwidth]{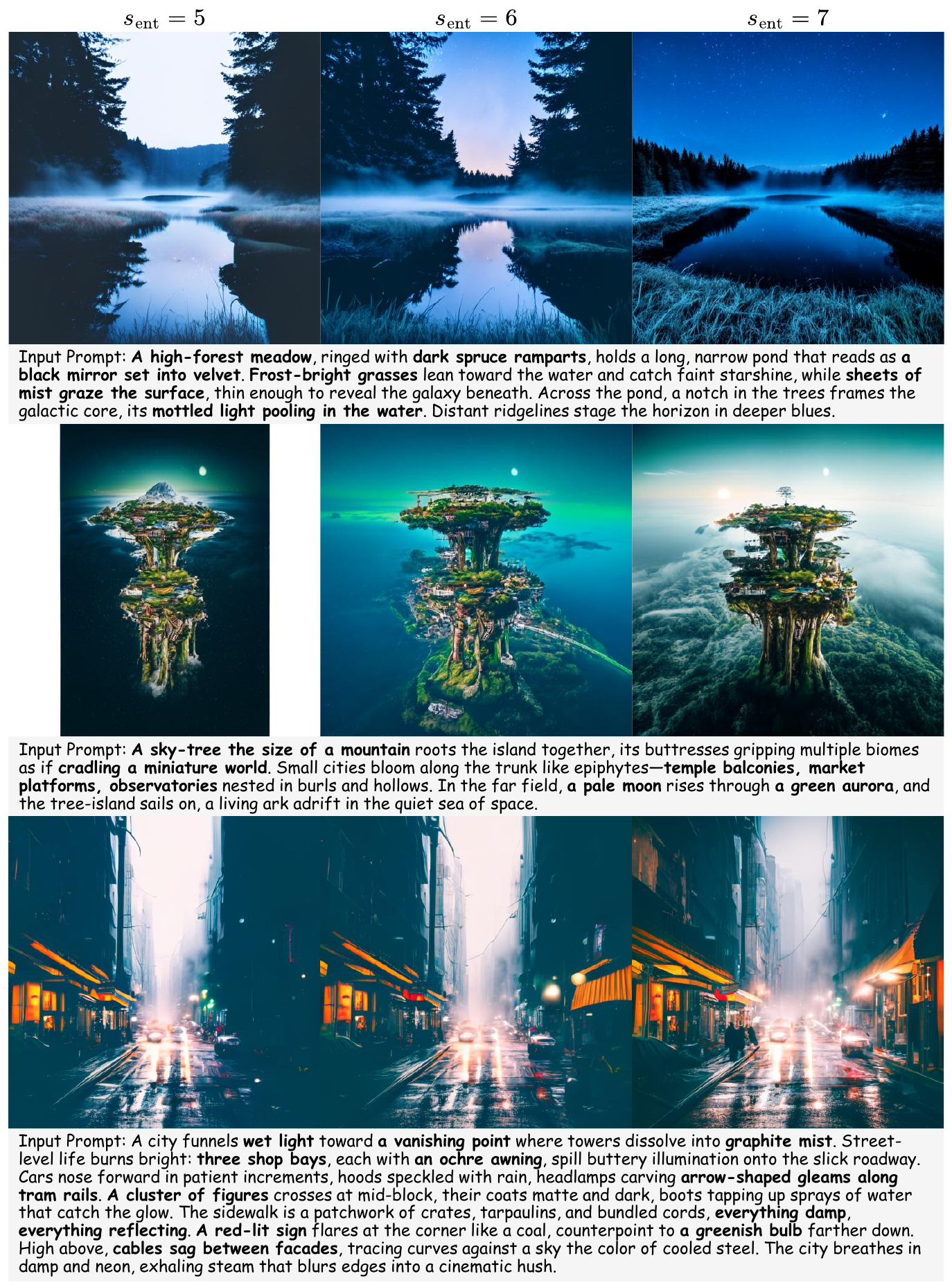}
    \caption{
        \textbf{Visualization comparison of images conditioned on entropy score $s_{\text{ent}}$}. From left to right, the columns correspond to $s_{\text{ent}}=5$, $6$, $7$, respectively.
    }
    \label{fig: token_ent}
\end{figure*}

\begin{figure*}[htbp]
    \centering
    \includegraphics[width=0.9\textwidth]{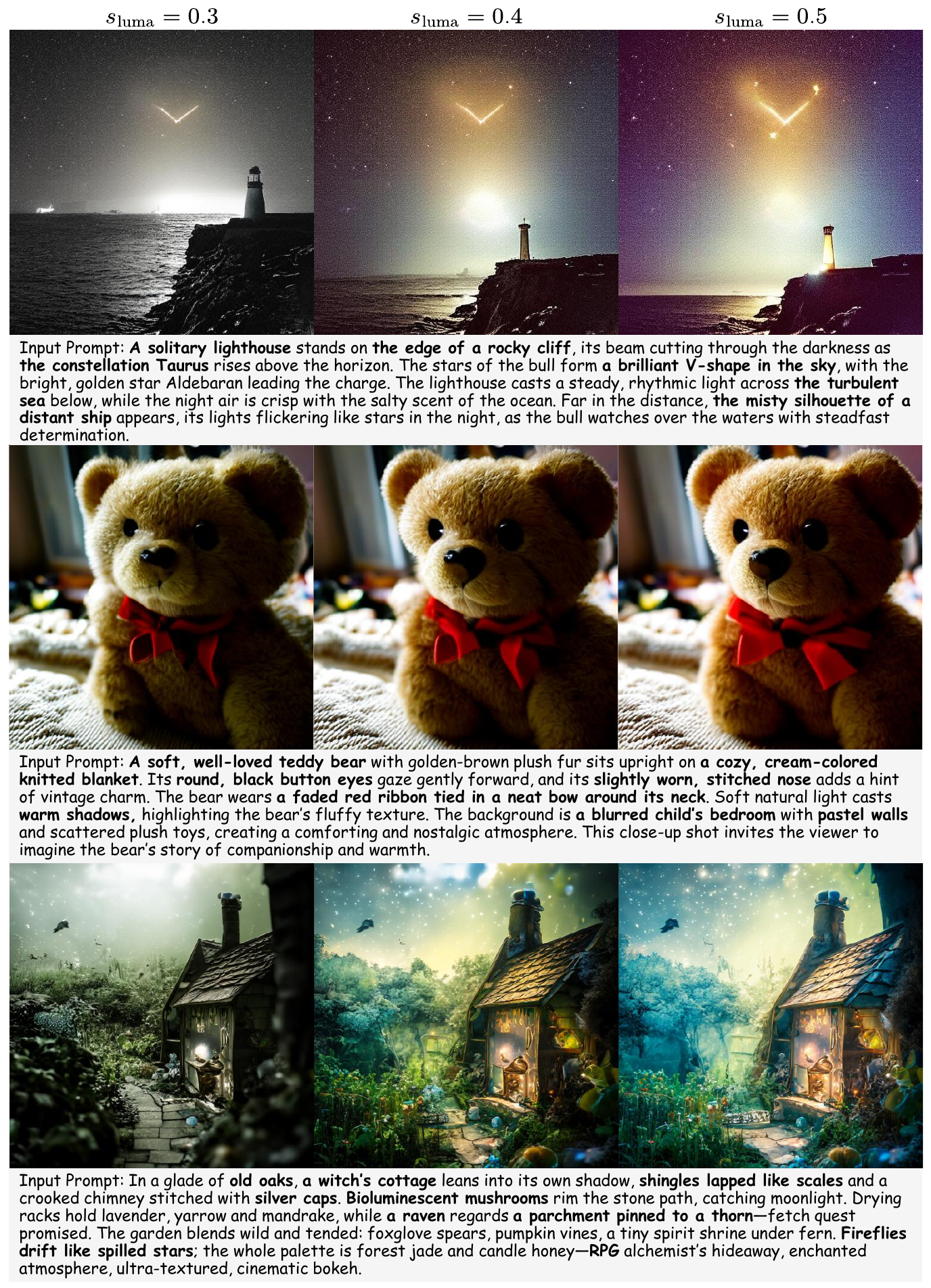}
    \caption{
        \textbf{Visualization comparison of images conditioned on luminance score $s_{\text{luma}}$}. From left to right, the columns correspond to $s_{\text{luma}}=0.3$, $0.4$, $0.5$, respectively.
    }
    \label{fig: token_luma}
\end{figure*}

\FloatBarrier

\section{Visualization Comparison}

\begin{figure*}[htbp]
    \centering
    \includegraphics[width=1.0\textwidth]{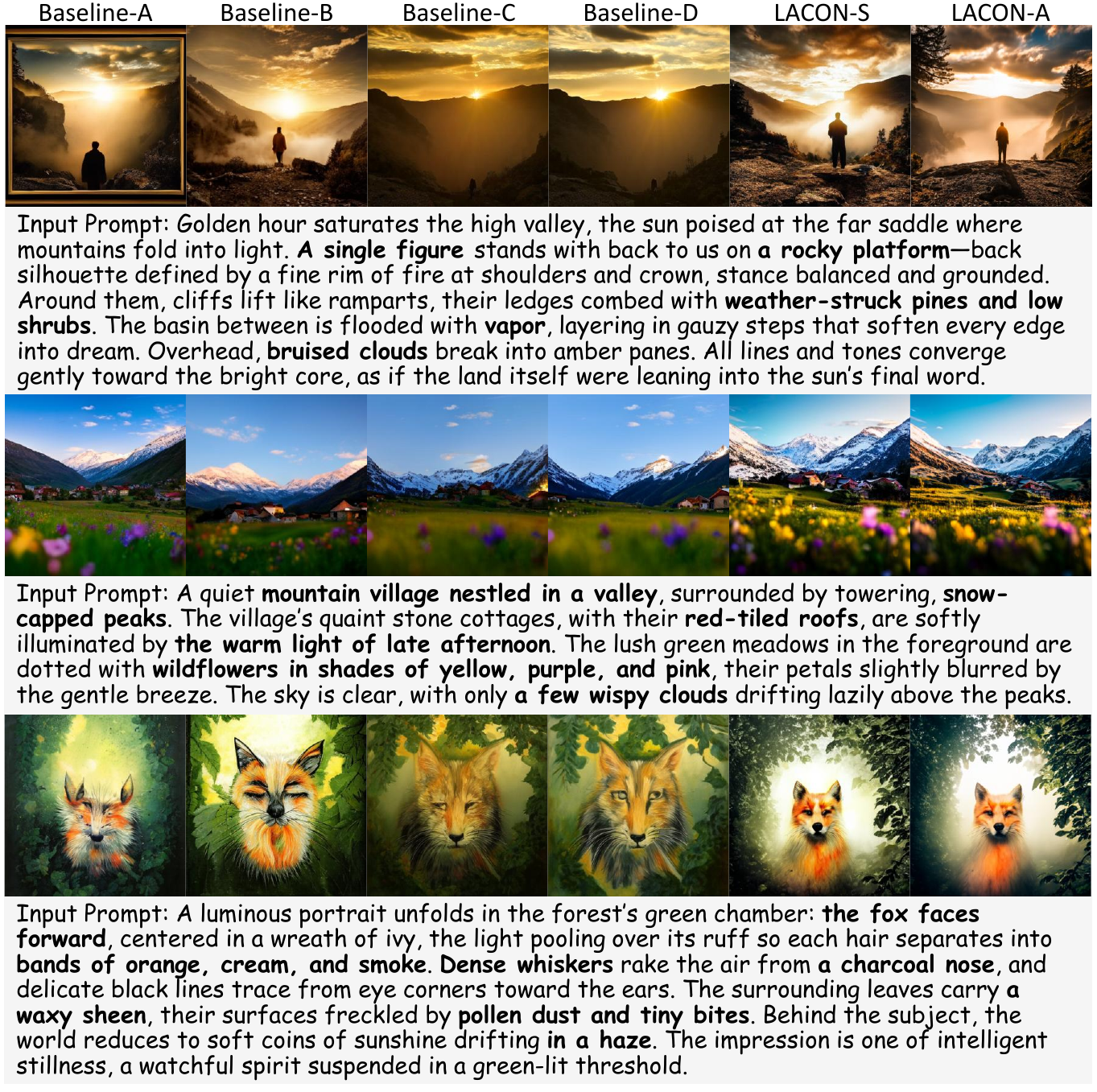}
    \caption{
        More visualization comparison of LACON-S and LACON-A against Baselines, demonstrating our LACON can still achieve superior visual generation quality even when trained on the full set of unfiltered raw images.
    }
    \label{fig: main_vis_part1}
\end{figure*}

\begin{figure*}[htbp]
    \centering
    \includegraphics[width=1.0\textwidth]{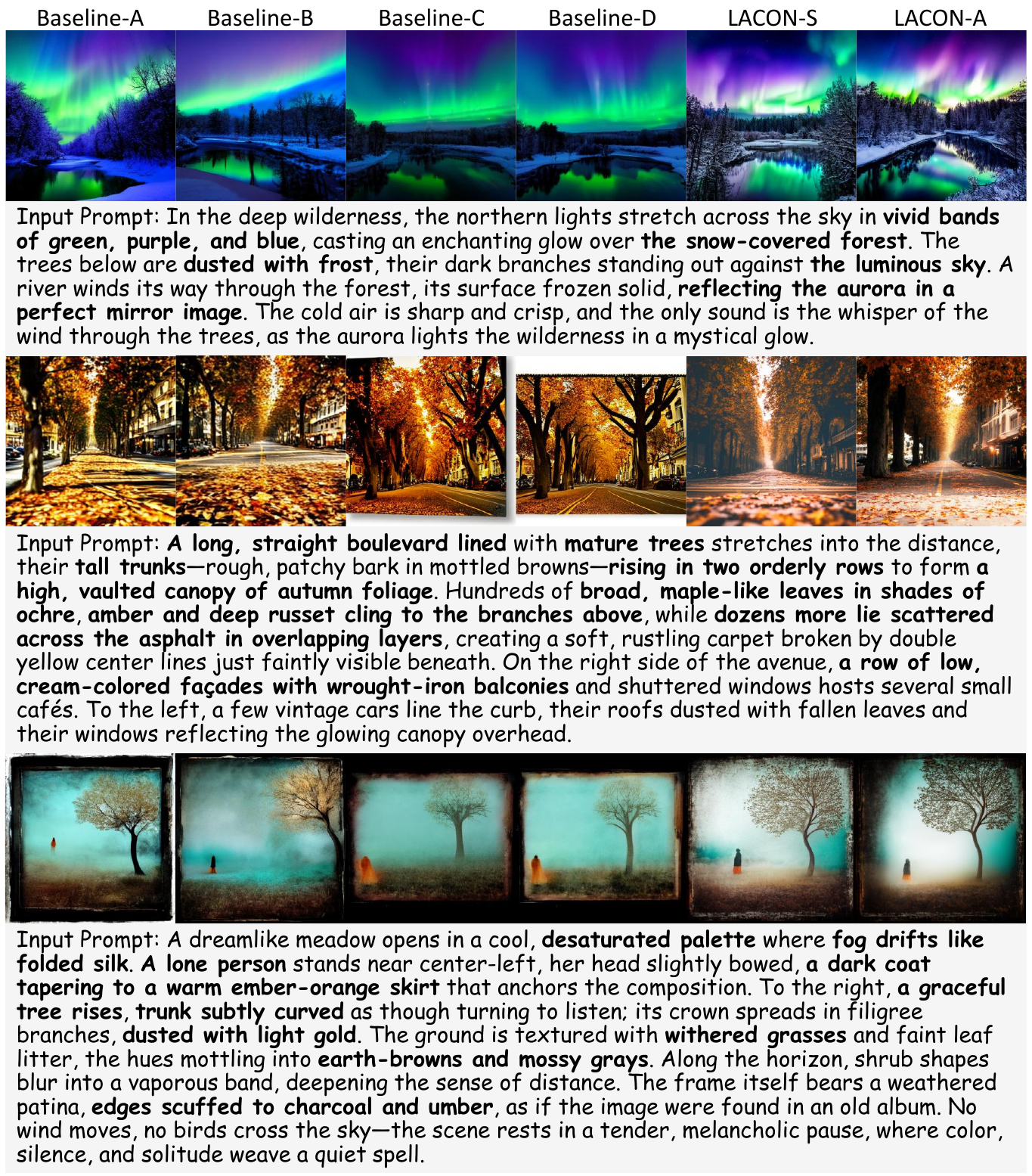}
    \caption{
        More visualization comparison of LACON-S and LACON-A against Baselines, demonstrating our LACON can still achieve superior visual generation quality even when trained on the full set of unfiltered raw images.
    }
    \label{fig: main_vis_part2}
\end{figure*}

\begin{figure*}[htbp]
    \centering
    \includegraphics[width=1.0\textwidth]{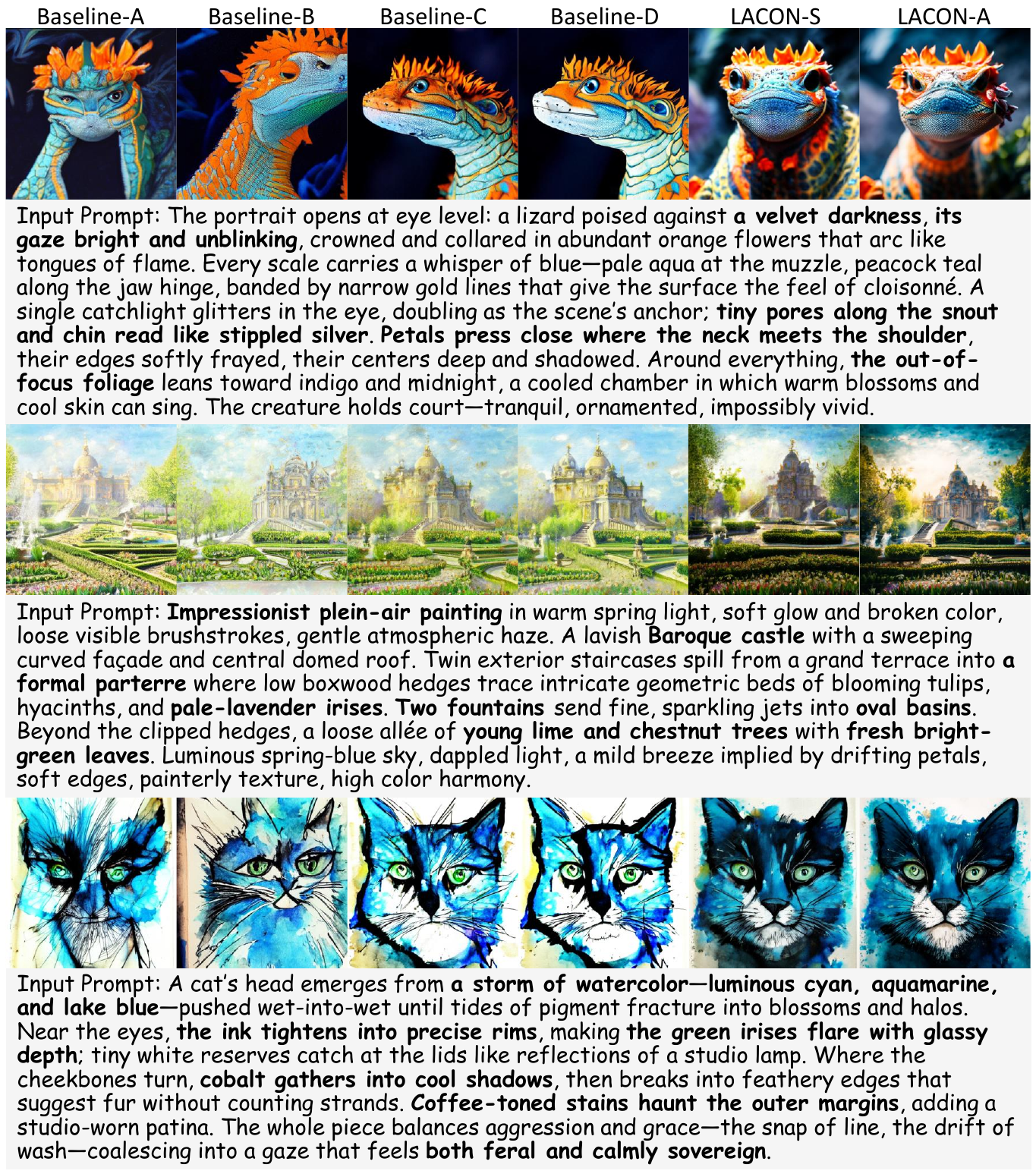}
    \caption{
        More visualization comparison of LACON-S and LACON-A against Baselines, demonstrating our LACON can still achieve superior visual generation quality even when trained on the full set of unfiltered raw images.
    }
    \label{fig: main_vis_part3}
\end{figure*}

\FloatBarrier

\section{Quality Signal Extraction Pipeline}
\label{sec:signals_acquire}
We build a simple yet effective pipeline to obtain quality signals for the full dataset. Specifically, the aesthetic score $s_{\text{aes}}$ is predicted using aesthetic-predictor-v2.5~\cite{schuhmann2022laion}, while the watermark score $s_{\text{wat}}$ is predicted using SigLIP2~\cite{tschannen2025siglip} fine-tuned for watermark detection. We further compute three no-reference image statistics. The clarity score $s_{\text{cla}}$ is defined as the variance of the Laplacian of a scale-normalized grayscale image and serves as a proxy for perceptual sharpness. The entropy score $s_{\text{ent}}$ is computed as the Shannon entropy of the grayscale intensity histogram, measuring information density. In addition, the luminance score $s_{\text{luma}}$ is calculated as the mean of the $V$ channel in the HSV color space (with values normalized to $[0,1]$), providing a no-reference estimate of visual brightness. These statistics are lightweight to compute and provide complementary views of image quality.

\begin{figure*}[htbp]
    \centering
    \includegraphics[width=0.6\textwidth]{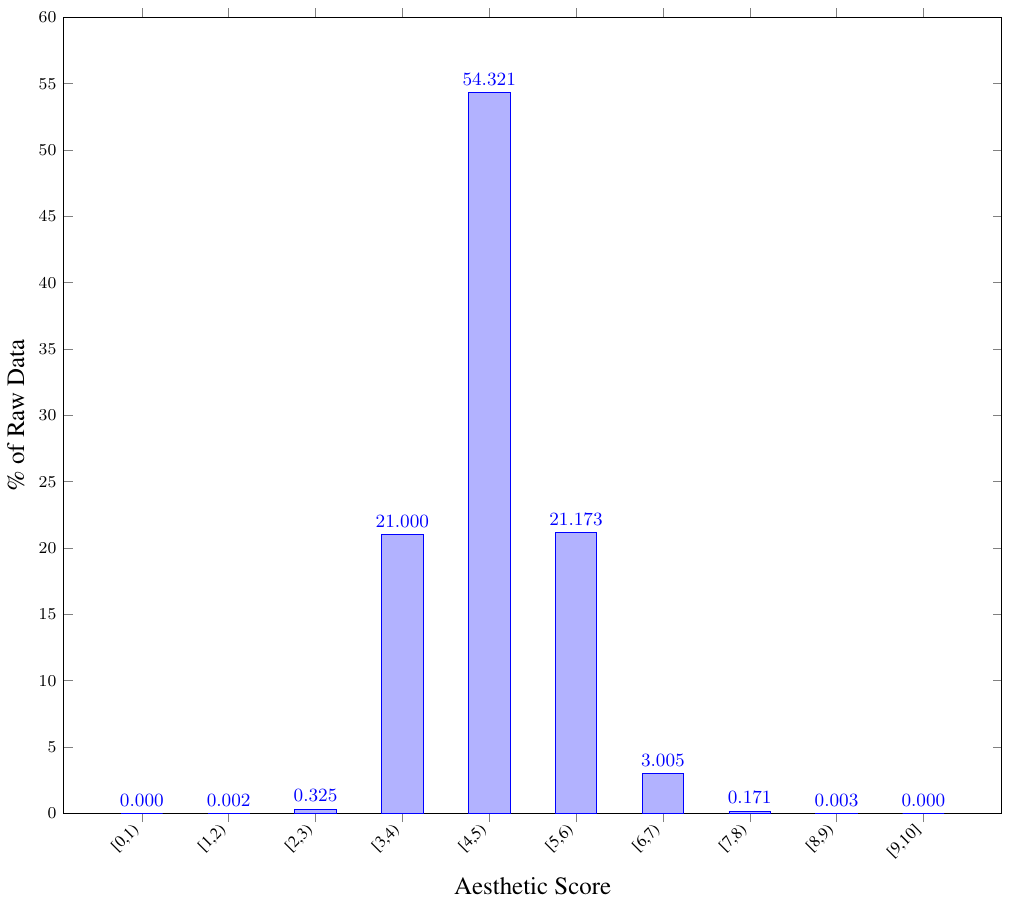}
    \caption{
        Aesthetic score distribution of the raw data. Each bar indicates the proportion of samples whose aesthetic score $s_{\text{aes}}$ falls within the corresponding score bin.
    }
    \label{fig: raw_aes}
\end{figure*}

\begin{figure*}[htbp]
    \centering
    \includegraphics[width=0.6\textwidth]{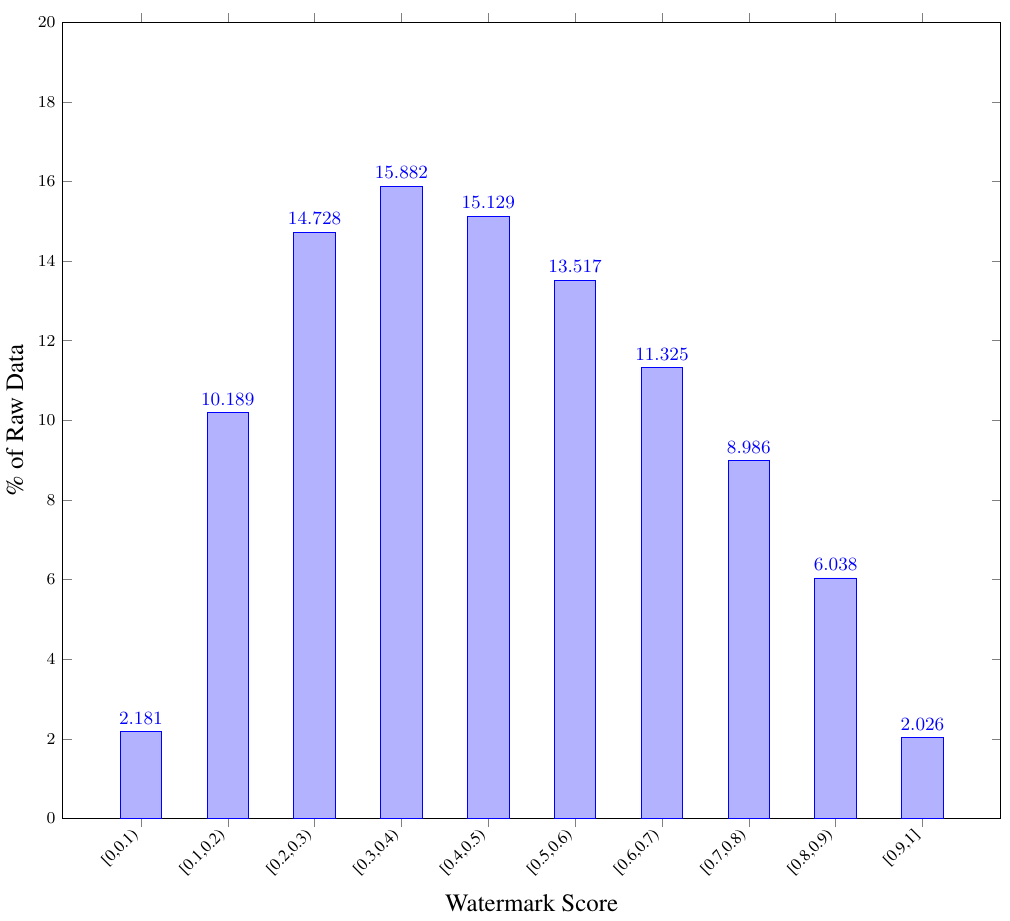}
    \caption{
        Watermark score distribution of the raw data. Each bar indicates the proportion of samples whose watermark score $s_{\text{wat}}$ falls within the corresponding score bin.
    }
    \label{fig: raw_wat}
\end{figure*}

\begin{figure*}[htbp]
    \centering
    \includegraphics[width=0.6\textwidth]{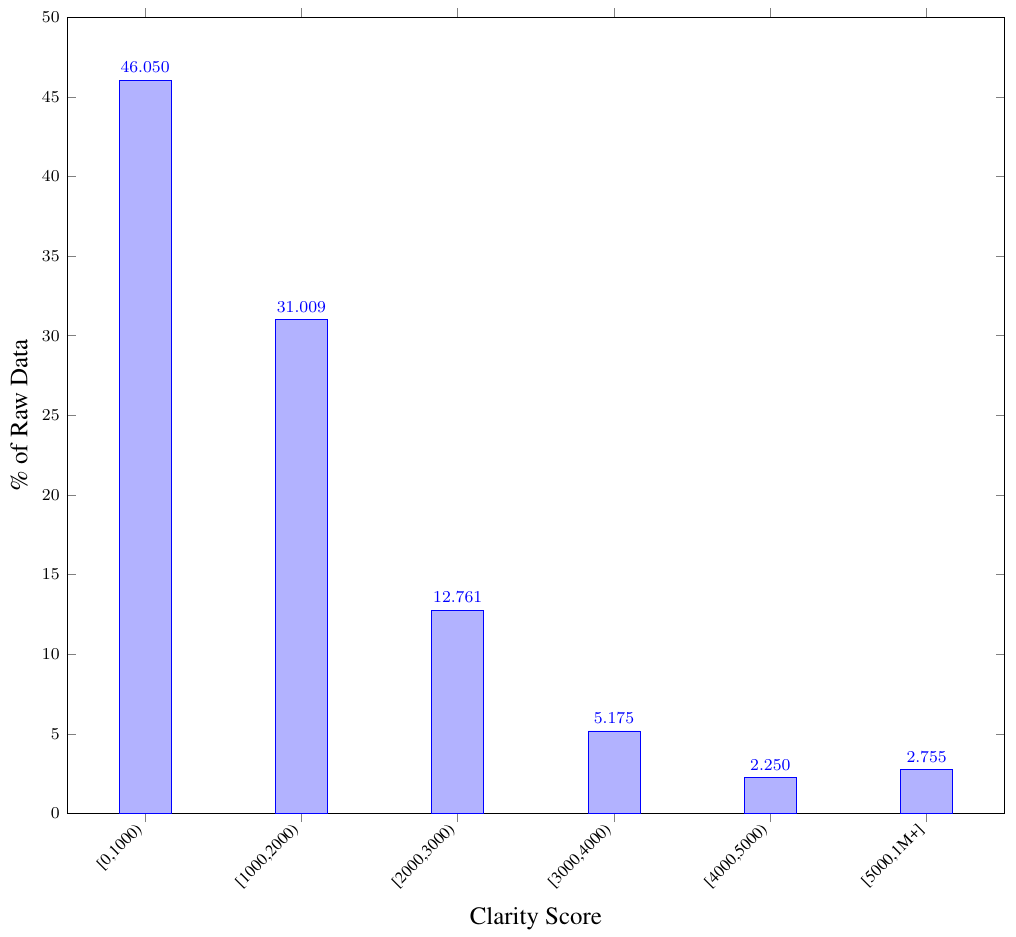}
    \caption{
        Clarity score distribution of the raw data. Each bar indicates the proportion of samples whose clarity score $s_{\text{cla}}$ falls within the corresponding score bin.
    }
    \label{fig: raw_cla}
\end{figure*}

\begin{figure*}[htbp]
    \centering
    \includegraphics[width=0.6\textwidth]{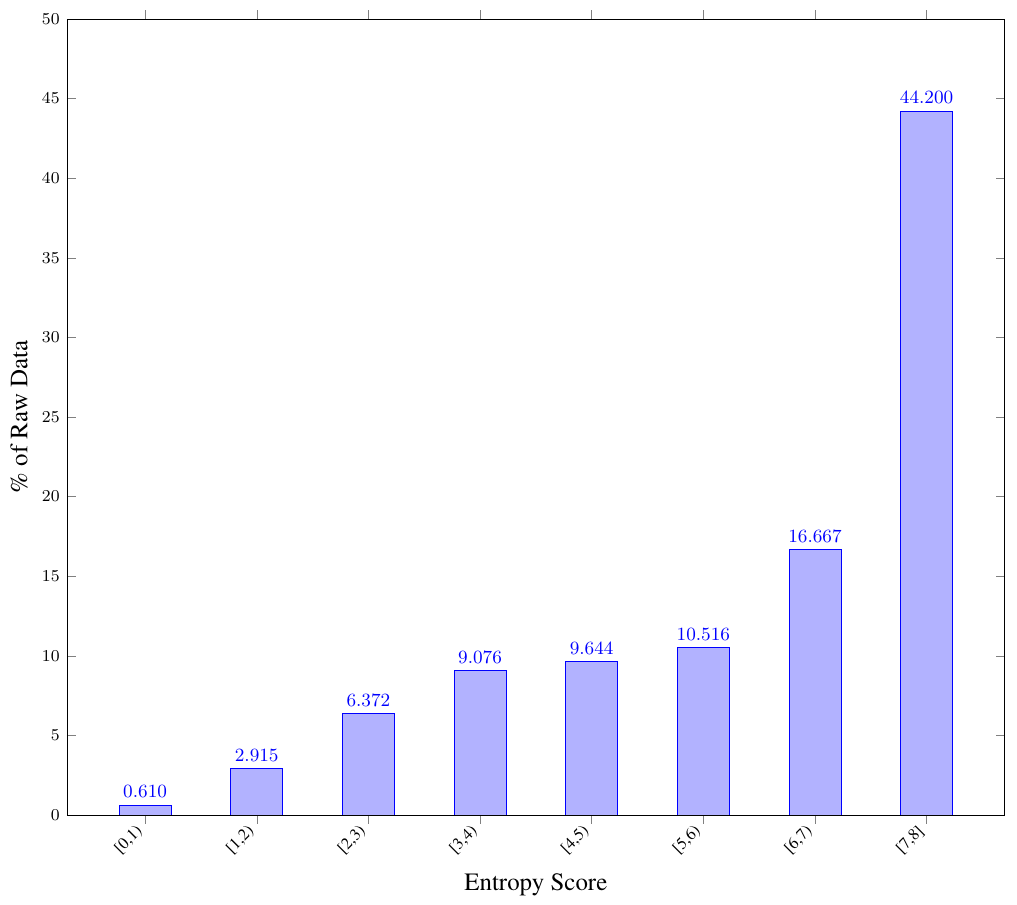}
    \caption{
        Entropy score distribution of the raw data. Each bar indicates the proportion of samples whose entropy score $s_{\text{ent}}$ falls within the corresponding score bin.
    }
    \label{fig: raw_ent}
\end{figure*}

\begin{figure*}[htbp]
    \centering
    \includegraphics[width=0.6\textwidth]{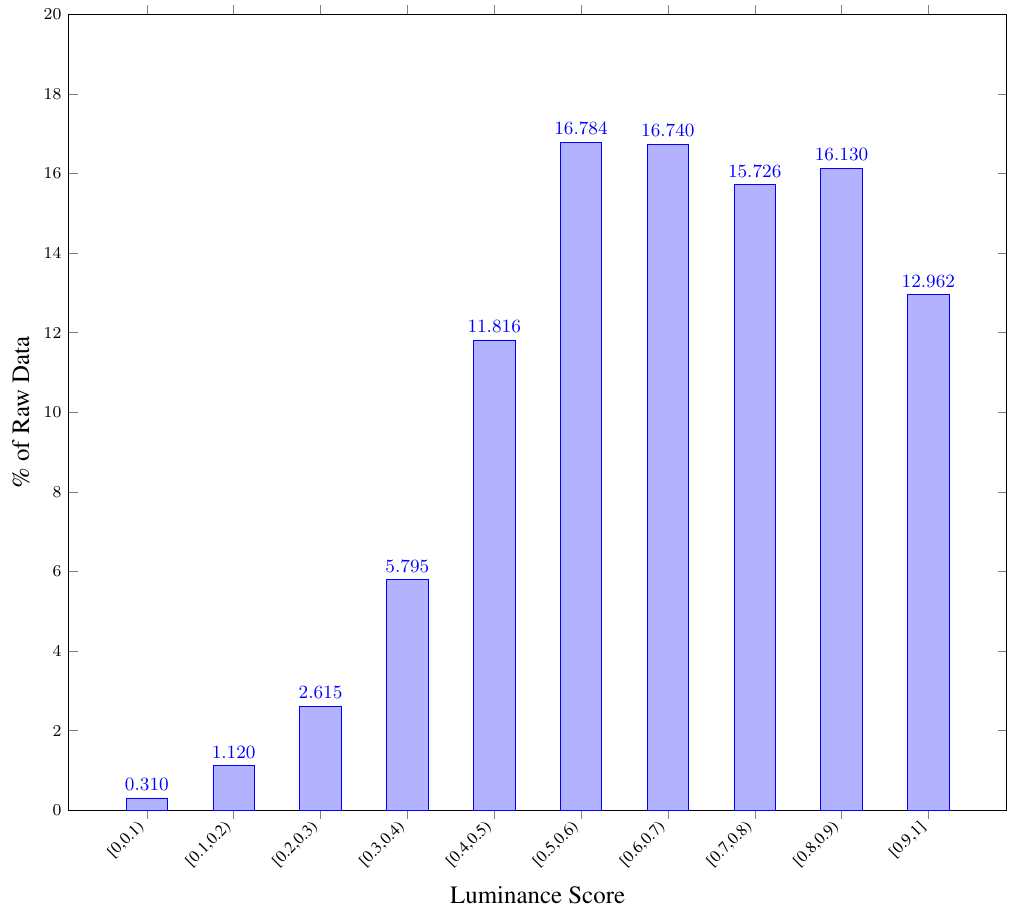}
    \caption{
        Luminance score distribution of the raw data. Each bar indicates the proportion of samples whose luminance score $s_{\text{luma}}$ falls within the corresponding score bin.
    }
    \label{fig: raw_luma}
\end{figure*}

The pipeline is fully automated and scalable, enabling us efficiently to compute quality signals for the full dataset under a consistent protocol. All scores are computed offline and stored alongside the data. The distributions of the resulting quality signals are illustrated in Figures~\ref{fig: raw_aes}--\ref{fig: raw_luma}.

\FloatBarrier

\section{Full Evaluation Results for Main Experiment}
The overall results for main experiment on GenEval and DPG are reported in the main text. Here, we provide a finer-grained breakdown by subcategory. Tables~\ref{tab:sana1.6B_ratio_geneval_detailed}–\ref{tab:sana1.6B_ratio_dpg_detailed} report detailed GenEval and DPG results for models trained on subsets obtained under different filtering thresholds. Tables~\ref{tab:sana0.6B_geneval_detailed}–\ref{tab:sana1.6B_dpg_detailed} present subcategory results comparing LACON with baseline methods on Sana-0.6B and Sana-1.6B. Finally, Tables~\ref{tab:sana1.6B_geneval_high_detailed}–\ref{tab:sana1.6B_dpg_high_detailed} provide the corresponding high-resolution results for Sana-1.6B, comparing LACON against the strongest baseline.

\begin{table*}[htbp]
  \centering
  \small
  \setlength{\tabcolsep}{4pt}
  \renewcommand{\arraystretch}{1.15}
  \caption{\textbf{Quantitative comparison on GenEval with details for Sana-1.6B trained on subsets obtained under different quality-signal filtering thresholds}. The results reveal a clear trade-off between data quantity and data quality in terms of model performance. We highlight the \textbf{best} and \underline{second-best} entries.}
  \label{tab:sana1.6B_ratio_geneval_detailed}

  \newcommand{\Hrule}{\noalign{\hrule height 1.2pt}}
  \newcommand{\Mrule}{\noalign{\hrule height 0.8pt}}
  \newcommand{\Srule}{\noalign{\hrule height 0.6pt}}

  \begin{tabular*}{\textwidth}{@{\extracolsep{\fill}} c|ccccccc}
    \specialrule{1.2pt}{0pt}{0pt}

    \multirow{2}{*}{\textbf{Model}}
      & \multirow{2}{*}{\textbf{Overall}~$\uparrow$}
      & \multicolumn{2}{c}{\textbf{Objects}}
      & \multirow{2}{*}{\textbf{Counting}}
      & \multirow{2}{*}{\textbf{Position}}
      & \multirow{2}{*}{\textbf{Colors}}
      & \multirow{2}{*}{\shortstack{\textbf{Color}\\\textbf{Attribution}}} \\
    \cline{3-4}
      & & \textbf{Single} & \textbf{Two} & & & & \\
    \specialrule{0.6pt}{0pt}{0pt}

    Ratio-5   & 58.3 & 92.2 & 55.8 & 47.8 & 43.8 & 75.5 & 34.8 \\
    Ratio-30  & 62.1 & 95.6 & 65.2 & \underline{50.3} & 42.5 & 78.2 & 41.3 \\
    Ratio-50  & 64.5 & \underline{97.8} & 69.7 & 44.1 & 50.3 & 77.7 & \underline{47.5} \\
    Ratio-65  & \textbf{68.0} & \underline{97.8} & \underline{72.7} & 48.1 & \textbf{57.0} & \textbf{82.2} & \textbf{50.0} \\
    Ratio-80  & \underline{67.5} & \textbf{98.1} & \textbf{73.2} & \textbf{53.4} & 53.0 & 80.9 & 46.5 \\
    Ratio-100 & 67.0 & 96.9 & 71.2 & \underline{50.3} & \underline{55.3} & \underline{81.4} & 46.8 \\

    \specialrule{1.2pt}{0pt}{0pt}
  \end{tabular*}
\end{table*}

\begin{table*}[htbp]
  \centering
  \small
  \setlength{\tabcolsep}{4pt}
  \renewcommand{\arraystretch}{1.15}
  \caption{\textbf{Quantitative comparison on DPG with details for Sana-1.6B trained on subsets obtained under different quality-signal filtering thresholds}. The results reveal a clear trade-off between data quantity and data quality in terms of model performance. We highlight the \textbf{best} and \underline{second-best} entries.}
  \label{tab:sana1.6B_ratio_dpg_detailed}

  \newcommand{\Hrule}{\noalign{\hrule height 1.2pt}}
  \newcommand{\Srule}{\noalign{\hrule height 0.6pt}}

  \begin{tabular*}{\textwidth}{@{\extracolsep{\fill}} c|cccccc}
    \Hrule
    \textbf{Model} & \textbf{Overall}~$\uparrow$ & Entity & Relation & Global & Attribute & Other \\
    \Srule
    Ratio-5   & 73.4 & 81.9 & 91.1 & 81.8 & 85.3 & 69.2 \\
    Ratio-30  & 73.6 & 82.5 & 91.0 & 81.5 & 85.6 & 67.2 \\
    Ratio-50  & 74.5 & 82.9 & \underline{91.7} & 81.8 & \textbf{85.9} & 64.4 \\
    Ratio-65  & \textbf{76.1} & \textbf{84.7} & \underline{91.7} & \textbf{82.1} & \underline{85.7} & \underline{70.8} \\
    Ratio-80  & \underline{75.8} & 83.7 & 91.3 & 79.6 & 85.4 & \textbf{71.6} \\
    Ratio-100 & 75.4 & \underline{83.8} & \textbf{92.0} & \textbf{82.1} & 85.6 & 68.0 \\
    \Hrule
  \end{tabular*}
\end{table*}

\begin{table*}[htbp]
  \centering
  \small
  \setlength{\tabcolsep}{4pt}
  \renewcommand{\arraystretch}{1.15}
  \caption{\textbf{Quantitative comparison between LACON and baseline methods for Sana-0.6B on GenEval with details}. Training on the full unfiltered dataset with explicit conditioning enables LACON to outperform both filter-first baselines and unconditioned baselines. We highlight the \textbf{best} and \underline{second-best} entries.}
  \label{tab:sana0.6B_geneval_detailed}

  \newcommand{\Hrule}{\noalign{\hrule height 1.2pt}}
  \newcommand{\Srule}{\noalign{\hrule height 0.6pt}}

  \begin{tabular*}{\textwidth}{@{\extracolsep{\fill}} c|ccccccc}
    \Hrule
    \multirow{2}{*}{\textbf{Model}}
      & \multirow{2}{*}{\textbf{Overall}~$\uparrow$}
      & \multicolumn{2}{c}{\textbf{Objects}}
      & \multirow{2}{*}{\textbf{Counting}}
      & \multirow{2}{*}{\textbf{Position}}
      & \multirow{2}{*}{\textbf{Colors}}
      & \multirow{2}{*}{\shortstack{\textbf{Color}\\\textbf{Attribution}}} \\
    \cline{3-4}
      & & \textbf{Single} & \textbf{Two} & & & & \\
    \Srule

    Baseline-A & 55.0 & 92.8 & 53.3 & 42.8 & 37.0 & 70.7 & 33.3 \\
    Baseline-B & \underline{62.8} & \underline{98.8} & 70.5 & 44.4 & 42.3 & \underline{82.2} & 38.8 \\
    Baseline-C & 60.8 & 96.6 & 64.7 & \underline{45.0} & 43.8 & 80.3 & 34.3 \\
    Baseline-D & 61.0 & 97.2 & 64.4 & 41.6 & 45.5 & 78.5 & 39.0 \\
    \noalign{\hrule height 0.8pt}
    \textbf{LACON-S}~(Ours) & \underline{64.4} & 98.1 & \underline{71.0} & 44.1 & \underline{46.8} & 81.9 & \textbf{45.0} \\
    \textbf{LACON-A}~(Ours) & \textbf{65.6} & \textbf{99.1} & \textbf{73.2} & \textbf{45.6} & \textbf{50.3} & \textbf{84.0} & \underline{41.3} \\
    \Hrule
  \end{tabular*}
\end{table*}

\begin{table*}[htbp]
  \centering
  \small
  \setlength{\tabcolsep}{4pt}
  \renewcommand{\arraystretch}{1.15}
  \caption{\textbf{Quantitative comparison between LACON and baseline methods for Sana-0.6B on DPG with details}. Training on the full unfiltered dataset with explicit conditioning enables LACON to outperform both filter-first baselines and unconditioned baselines. We highlight the \textbf{best} and \underline{second-best} entries.}
  \label{tab:sana0.6B_dpg_detailed}

  \newcommand{\Hrule}{\noalign{\hrule height 1.2pt}}
  \newcommand{\Srule}{\noalign{\hrule height 0.6pt}}

  \begin{tabular*}{\textwidth}{@{\extracolsep{\fill}} c|cccccc}
    \Hrule
    \textbf{Model} & \textbf{Overall}~$\uparrow$ & Entity & Relation & Global & Attribute & Other \\
    \Srule
    Baseline-A   & 68.1 & 79.0 & 89.5 & 77.2 & 84.8 & 59.6 \\
    Baseline-B  & 70.4 & 80.0 & 90.0 & \textbf{83.6} & 85.2 & 62.8 \\
    Baseline-C  & 69.7 & 79.8 & 90.4 & 80.9 & 84.9 & 62.4 \\
    Baseline-D  & 69.2 & 79.4 & 89.7 & \underline{81.8} & 84.2 & 60.8 \\
    \specialrule{0.8pt}{0pt}{0pt}
    \textbf{LACON-S}~(Ours)  & \textbf{71.9} & \underline{80.8} & \textbf{90.6} & 81.5 & \underline{85.4} & \underline{64.4} \\
    \textbf{LACON-A}~(Ours) & \underline{71.8} & \textbf{81.1} & \underline{90.5} & 81.2 & \textbf{85.5} & \textbf{65.2} \\
    \Hrule
  \end{tabular*}
\end{table*}

\begin{table*}[htbp]
  \centering
  \small
  \setlength{\tabcolsep}{4pt}
  \renewcommand{\arraystretch}{1.15}
  \caption{\textbf{Quantitative comparison between LACON and baseline methods for Sana-1.6B on GenEval with details}. Training on the full unfiltered dataset with explicit conditioning enables LACON to outperform both filter-first baselines and unconditioned baselines. We highlight the \textbf{best} and \underline{second-best} entries.}
  \label{tab:sana1.6B_geneval_detailed}

  \newcommand{\Hrule}{\noalign{\hrule height 1.2pt}}
  \newcommand{\Srule}{\noalign{\hrule height 0.6pt}}

  \begin{tabular*}{\textwidth}{@{\extracolsep{\fill}} c|ccccccc}
    \Hrule
    \multirow{2}{*}{\textbf{Model}}
      & \multirow{2}{*}{\textbf{Overall}~$\uparrow$}
      & \multicolumn{2}{c}{\textbf{Objects}}
      & \multirow{2}{*}{\textbf{Counting}}
      & \multirow{2}{*}{\textbf{Position}}
      & \multirow{2}{*}{\textbf{Colors}}
      & \multirow{2}{*}{\shortstack{\textbf{Color}\\\textbf{Attribution}}} \\
    \cline{3-4}
      & & \textbf{Single} & \textbf{Two} & & & & \\
    \Srule

    Baseline-A   & 58.3 & 92.2 & 55.8 & 47.8 & 43.8 & 75.5 & 34.8 \\
    Baseline-B   & 68.0 & 97.8 & 72.7 & 48.1 & \textbf{57.0} & 82.2 & \underline{50.0} \\
    Baseline-C   & 67.0 & 96.9 & 71.2 & 50.3 & 55.3 & 81.4 & 46.8 \\
    Baseline-D   & 67.1 & 95.3 & 68.4 & \textbf{59.7} & 51.5 & 81.9 & 45.5 \\
    \noalign{\hrule height 0.8pt}
    \textbf{LACON-S}~(Ours) & \underline{70.9} & \textbf{99.1} & \underline{78.8} & 56.9 & \underline{56.8} & \textbf{86.2} & 48.0 \\
    \textbf{LACON-A}~(Ours) & \textbf{71.6} & \underline{98.6} & \textbf{79.0} & \underline{57.8} & 56.3 & \underline{85.9} & \textbf{51.8} \\
    \Hrule
  \end{tabular*}
\end{table*}

\begin{table*}[htbp]
  \centering
  \small
  \setlength{\tabcolsep}{4pt}
  \renewcommand{\arraystretch}{1.15}
  \caption{\textbf{Quantitative comparison between LACON and baseline methods for Sana-1.6B on DPG with details}. Training on the full unfiltered dataset with explicit conditioning enables LACON to outperform both filter-first baselines and unconditioned baselines. We highlight the \textbf{best} and \underline{second-best} entries.}
  \label{tab:sana1.6B_dpg_detailed}

  \newcommand{\Hrule}{\noalign{\hrule height 1.2pt}}
  \newcommand{\Srule}{\noalign{\hrule height 0.6pt}}

  \begin{tabular*}{\textwidth}{@{\extracolsep{\fill}} l|cccccc}
    \Hrule
    \textbf{Model} & \textbf{Overall}~$\uparrow$ & Entity & Relation & Global & Attribute & Other \\
    \Srule
    Baseline-A   & 73.4 & 81.9 & 91.1 & 81.8 & 85.3 & 69.2 \\
    Baseline-B  & 76.1 & 84.7 & 91.7 & \textbf{82.1} & 85.7 & 70.8 \\
    Baseline-C  & 75.4 & 83.8 & 92.0 & \textbf{82.1} & 85.6 & 68.0 \\
    Baseline-D  & 75.5 & 84.1 & 91.2 & 80.9 & 85.3 & \underline{72.8} \\
    \specialrule{0.8pt}{0pt}{0pt}
    \textbf{LACON-S}~(Ours)  & \underline{77.3} & \underline{85.8} & \textbf{92.7} & 80.9 & \underline{86.8} & 70.8 \\
    \textbf{LACON-A}~(Ours) & \textbf{78.1} & \textbf{86.1} & \underline{92.6} & \textbf{82.1} & \textbf{87.9} & \textbf{76.0} \\
    \Hrule
  \end{tabular*}
\end{table*}

\begin{table*}[htbp]
  \centering
  \small
  \setlength{\tabcolsep}{4pt}
  \renewcommand{\arraystretch}{1.15}
  \caption{\textbf{Quantitative comparison between LACON and the strongest baseline for Sana-1.6B at 1024$\times$1024 resolution on GenEval with details}. We highlight the \textbf{best} and \underline{second-best} entries.}
  \label{tab:sana1.6B_geneval_high_detailed}

  \newcommand{\Hrule}{\noalign{\hrule height 1.2pt}}
  \newcommand{\Srule}{\noalign{\hrule height 0.6pt}}

  \begin{tabular*}{\textwidth}{@{\extracolsep{\fill}} c|ccccccc}
    \Hrule
    \multirow{2}{*}{\textbf{Model}}
      & \multirow{2}{*}{\textbf{Overall}~$\uparrow$}
      & \multicolumn{2}{c}{\textbf{Objects}}
      & \multirow{2}{*}{\textbf{Counting}}
      & \multirow{2}{*}{\textbf{Position}}
      & \multirow{2}{*}{\textbf{Colors}}
      & \multirow{2}{*}{\shortstack{\textbf{Color}\\\textbf{Attribution}}} \\
    \cline{3-4}
      & & \textbf{Single} & \textbf{Two} & & & & \\
    \Srule
    Baseline-B   & 68.3 & 97.5 & \underline{77.5} & \underline{54.1} & 54.5 & 82.5 & 43.5 \\
    \noalign{\hrule height 0.8pt}
    \textbf{LACON-S}~(Ours) & \underline{70.9} & \textbf{98.8} & 76.3 & \underline{54.1} & \textbf{59.5} & \underline{85.1} & \textbf{52.0} \\
    \textbf{LACON-A}~(Ours) & \textbf{71.5} & \textbf{98.8} & \textbf{80.1} & \textbf{57.5} & \underline{55.5} & \textbf{86.7} & \underline{50.8} \\
    \Hrule
  \end{tabular*}
\end{table*}

\clearpage
\begin{table*}[htbp]
  \centering
  \small
  \setlength{\tabcolsep}{4pt}
  \renewcommand{\arraystretch}{1.15}
  \caption{\textbf{Quantitative comparison between LACON and the strongest baseline for Sana-1.6B at 1024$\times$1024 resolution on DPG with details}. We highlight the \textbf{best} and \underline{second-best} entries.}
  \label{tab:sana1.6B_dpg_high_detailed}

  \newcommand{\Hrule}{\noalign{\hrule height 1.2pt}}
  \newcommand{\Srule}{\noalign{\hrule height 0.6pt}}

  \begin{tabular*}{\textwidth}{@{\extracolsep{\fill}} l|cccccc}
    \Hrule
    \textbf{Model} & \textbf{Overall}~$\uparrow$ & Entity & Relation & Global & Attribute & Other \\
    \Srule
    Baseline-B  & 76.3 & 84.9 & 91.5 & 80.9 & 86.6 & 70.0 \\
    \specialrule{0.8pt}{0pt}{0pt}
    \textbf{LACON-S}~(Ours)  & \underline{77.6} & \textbf{86.2} & \underline{92.4} & \textbf{82.1} & \textbf{88.0} & \underline{74.8} \\
    \textbf{LACON-A}~(Ours) & \textbf{78.8} & \underline{85.9} & \textbf{93.0} & \textbf{82.1} & \underline{86.7} & \textbf{80.0} \\
    \Hrule
  \end{tabular*}
\end{table*}


\section{Full Evaluation Results for Autoregressive Architecture}

The overall results for autoregressive architecture on GenEval and DPG are reported in the main text. Here, we provide a finer-grained breakdown by subcategory. Tables~\ref{tab:qwen3_0.6B_geneval_detailed}–\ref{tab:qwen3_1.7B_dpg_detailed} report detailed GenEval and DPG results comparing LACON with the strongest baseline on Qwen3-0.6B and Qwen3-1.7B.

\begin{table*}[htbp]
  \centering
  \small
  \setlength{\tabcolsep}{4pt}
  \renewcommand{\arraystretch}{1.15}
  \caption{\textbf{Quantitative comparison between LACON and the strongest baseline for Qwen3-0.6B on GenEval with details}, indicating that LACON generalizes well to autoregressive architecture. We highlight the \textbf{best} and \underline{second-best} entries.}
  \label{tab:qwen3_0.6B_geneval_detailed}

  \newcommand{\Hrule}{\noalign{\hrule height 1.2pt}}
  \newcommand{\Srule}{\noalign{\hrule height 0.6pt}}

  \begin{tabular*}{\textwidth}{@{\extracolsep{\fill}} c|ccccccc}
    \Hrule
    \multirow{2}{*}{\textbf{Model}}
      & \multirow{2}{*}{\textbf{Overall}~$\uparrow$}
      & \multicolumn{2}{c}{\textbf{Objects}}
      & \multirow{2}{*}{\textbf{Counting}}
      & \multirow{2}{*}{\textbf{Position}}
      & \multirow{2}{*}{\textbf{Colors}}
      & \multirow{2}{*}{\shortstack{\textbf{Color}\\\textbf{Attribution}}} \\
    \cline{3-4}
      & & \textbf{Single} & \textbf{Two} & & & & \\
    \Srule
    Baseline-B   & 58.9 & \textbf{99.4} & \underline{73.5} & 26.3 & 46.5 & \textbf{77.9} & 30.0 \\
    \noalign{\hrule height 0.8pt}
    \textbf{LACON-S}~(Ours) & \textbf{61.3} & \underline{98.8} & 70.7 & \textbf{39.1} & \underline{51.5} & 75.0 & \textbf{32.5} \\
    \textbf{LACON-A}~(Ours) & \textbf{61.3} & \underline{98.8} & \textbf{78.3} & \underline{28.1} & \textbf{54.0} & \underline{77.7} & \underline{31.0} \\
    \Hrule
  \end{tabular*}
\end{table*}

\begin{table*}[htbp]
  \centering
  \small
  \setlength{\tabcolsep}{4pt}
  \renewcommand{\arraystretch}{1.15}
  \caption{\textbf{Quantitative comparison between LACON and the strongest baseline for Qwen3-0.6B on DPG with details}, indicating that LACON generalizes well to autoregressive architecture. We highlight the \textbf{best} and \underline{second-best} entries.}
  \label{tab:qwen3_0.6B_dpg_detailed}

  \newcommand{\Hrule}{\noalign{\hrule height 1.2pt}}
  \newcommand{\Srule}{\noalign{\hrule height 0.6pt}}

  \begin{tabular*}{\textwidth}{@{\extracolsep{\fill}} c|cccccc}
    \Hrule
    \textbf{Model} & \textbf{Overall}~$\uparrow$ & Entity & Relation & Global & Attribute & Other \\
    \Srule
    Baseline-B  & 73.1 & 82.7 & \underline{91.4} & \underline{79.3} & 84.0 & 69.6 \\
    \specialrule{0.8pt}{0pt}{0pt}
    \textbf{LACON-S}~(Ours)  & \underline{74.9} & \underline{84.1} & \textbf{92.6} & \textbf{80.9} & \underline{84.6} & \underline{71.6} \\
    \textbf{LACON-A}~(Ours) & \textbf{75.4} & \textbf{84.3} & 91.0 & \underline{79.3} & \textbf{86.0} & \textbf{73.2} \\
    \Hrule
  \end{tabular*}
\end{table*}

\begin{table*}[htbp]
  \centering
  \small
  \setlength{\tabcolsep}{4pt}
  \renewcommand{\arraystretch}{1.15}
  \caption{\textbf{Quantitative comparison between LACON and the strongest baseline for Qwen3-1.7B on GenEval with details}, indicating that LACON generalizes well to autoregressive architecture. We highlight the \textbf{best} and \underline{second-best} entries.}
  \label{tab:qwen3_1.7B_geneval_detailed}

  \newcommand{\Hrule}{\noalign{\hrule height 1.2pt}}
  \newcommand{\Srule}{\noalign{\hrule height 0.6pt}}

  \begin{tabular*}{\textwidth}{@{\extracolsep{\fill}} c|ccccccc}
    \Hrule
    \multirow{2}{*}{\textbf{Model}}
      & \multirow{2}{*}{\textbf{Overall}~$\uparrow$}
      & \multicolumn{2}{c}{\textbf{Objects}}
      & \multirow{2}{*}{\textbf{Counting}}
      & \multirow{2}{*}{\textbf{Position}}
      & \multirow{2}{*}{\textbf{Colors}}
      & \multirow{2}{*}{\shortstack{\textbf{Color}\\\textbf{Attribution}}} \\
    \cline{3-4}
      & & \textbf{Single} & \textbf{Two} & & & & \\
    \Srule
    Baseline-B   & 66.4 & 98.1 & 82.8 & \underline{36.3} & 59.8 & 76.9 & 44.5 \\
    \noalign{\hrule height 0.8pt}
    \textbf{LACON-S}~(Ours) & \underline{69.7} & \underline{99.1} & \textbf{88.9} & 33.1 & \underline{60.8} & \textbf{86.2} & \textbf{50.5} \\
    \textbf{LACON-A}~(Ours) & \textbf{70.3} & \textbf{99.7} & \underline{84.1} & \textbf{42.2} & \textbf{62.0} & \underline{84.3} & \underline{49.8} \\
    \Hrule
  \end{tabular*}
\end{table*}

\begin{table*}[htbp]
  \centering
  \small
  \setlength{\tabcolsep}{4pt}
  \renewcommand{\arraystretch}{1.15}
  \caption{\textbf{Quantitative comparison between LACON and the strongest baseline for Qwen3-1.7B on DPG with details}, indicating that LACON generalizes well to autoregressive architecture. We highlight the \textbf{best} and \underline{second-best} entries.}
  \label{tab:qwen3_1.7B_dpg_detailed}

  \newcommand{\Hrule}{\noalign{\hrule height 1.2pt}}
  \newcommand{\Srule}{\noalign{\hrule height 0.6pt}}

  \begin{tabular*}{\textwidth}{@{\extracolsep{\fill}} c|cccccc}
    \Hrule
    \textbf{Model} & \textbf{Overall}~$\uparrow$ & Entity & Relation & Global & Attribute & Other \\
    \Srule
    Baseline-B  & 78.2 & 86.3 & \textbf{92.6} & \textbf{81.5} & 85.1 & 72.0 \\
    \specialrule{0.8pt}{0pt}{0pt}
    \textbf{LACON-S}~(Ours)  & \underline{79.4} & \underline{86.9} & 92.1 & \underline{81.4} & \textbf{86.7} & \textbf{74.8} \\
    \textbf{LACON-A}~(Ours) & \textbf{80.1} & \textbf{87.0} & \underline{92.4} & 80.2 & \textbf{86.7} & \underline{72.8} \\
    \Hrule
  \end{tabular*}
\end{table*}

\FloatBarrier
\section{Full Evaluation Results for Token-injection Strategies}

The overall results for token-injection strategies on the GenEval and DPG are reported in the main text. Here, we provide a finer-grained breakdown by subcategory. Tables~\ref{tab:sana1.6B_strategies_geneval_detailed}–\ref{tab:sana1.6B_strategies_dpg_detailed} report detailed GenEval and DPG results for Sana-1.6B across different token-injection strategies.

\begin{table*}[htbp]
  \centering
  \small
  \setlength{\tabcolsep}{4pt}
  \renewcommand{\arraystretch}{1.15}
  \caption{\textbf{Quantitative comparison of different token-injection strategies for Sana-1.6B on GenEval with details}. The results indicate that our GCC token-injection strategy can achieve better performance on most subcategories. We highlight the \textbf{best} and \underline{second-best} entries.}
  \label{tab:sana1.6B_strategies_geneval_detailed}

  \newcommand{\Hrule}{\noalign{\hrule height 1.2pt}}
  \newcommand{\Srule}{\noalign{\hrule height 0.6pt}}

  \begin{tabular*}{\textwidth}{@{\extracolsep{\fill}} c|ccccccc}
    \Hrule
    \multirow{2}{*}{\textbf{Model}}
      & \multirow{2}{*}{\textbf{Overall}~$\uparrow$}
      & \multicolumn{2}{c}{\textbf{Objects}}
      & \multirow{2}{*}{\textbf{Counting}}
      & \multirow{2}{*}{\textbf{Position}}
      & \multirow{2}{*}{\textbf{Colors}}
      & \multirow{2}{*}{\shortstack{\textbf{Color}\\\textbf{Attribution}}} \\
    \cline{3-4}
      & & \textbf{Single} & \textbf{Two} & & & & \\
    \Srule
    Linear Interpolation   & 69.5 & 97.8 & 73.5 & 57.2 & 54.3 & 84.3 & \underline{50.3} \\
    Discrete Binning   & 69.8 & 98.1 & 74.5 & 56.3 & \textbf{59.8} & 83.2 & \underline{50.3} \\
    Fourier Feature   & \underline{70.4} & \textbf{99.1} & \underline{77.3} & \textbf{57.8} & 53.3 & \textbf{85.9} & 47.0 \\
    \noalign{\hrule height 0.8pt}
    \textbf{GCC}~(Ours) & \textbf{71.6} & \underline{98.6} & \textbf{79.0} & \textbf{57.8} & \underline{56.3} & \textbf{85.9} & \textbf{51.8} \\
    \Hrule
  \end{tabular*}
\end{table*}

\begin{table*}[htbp]
  \centering
  \small
  \setlength{\tabcolsep}{4pt}
  \renewcommand{\arraystretch}{1.15}
  \caption{\textbf{Quantitative comparison of different token-injection strategies for Sana-1.6B on DPG with details}. The results indicate that our GCC token-injection strategy can achieve better performance on most subcategories. We highlight the \textbf{best} and \underline{second-best} entries.}
  \label{tab:sana1.6B_strategies_dpg_detailed}

  \newcommand{\Hrule}{\noalign{\hrule height 1.2pt}}
  \newcommand{\Srule}{\noalign{\hrule height 0.6pt}}

  \begin{tabular*}{\textwidth}{@{\extracolsep{\fill}} c|cccccc}
    \Hrule
    \textbf{Model} & \textbf{Overall}~$\uparrow$ & Entity & Relation & Global & Attribute & Other \\
    \Srule
    Linear Interpolation  & 76.1 & \underline{85.0} & 92.5 & 80.2 & \underline{86.4} & 71.6 \\
    Discrete Binning  & \underline{77.2} & 84.8 & \textbf{92.6} & 81.2 & 86.2 & 72.4 \\
    Fourier Feature  & 77.0 & 84.1 & 91.7 & \textbf{82.1} & 86.3 & \underline{74.0} \\
    \specialrule{0.8pt}{0pt}{0pt}
    \textbf{GCC}~(Ours) & \textbf{78.1} & \textbf{86.1} & \textbf{92.6} & \textbf{82.1} & \textbf{87.9} & \textbf{76.0} \\
    \Hrule
  \end{tabular*}
\end{table*}

\FloatBarrier

\end{document}